\pdfoutput=1
\documentclass[11pt]{article}

\usepackage{naacl2021}

\usepackage{times}
\usepackage{latexsym}
\usepackage{booktabs}
\usepackage{xcolor}
\usepackage{latexsym}
\usepackage{graphicx,multirow}
\usepackage{subcaption}

\usepackage{subfiles}

\usepackage{natbib}

\usepackage[T1]{fontenc}
\usepackage[utf8]{inputenc}

\usepackage{microtype}

\title{Multilingual Language Models Predict Human Reading Behavior}

\author{Nora Hollenstein\textsuperscript{1}, Federico Pirovano\textsuperscript{1}, Ce Zhang\textsuperscript{1}, Lena Jäger\textsuperscript{2,3}, Lisa Beinborn\textsuperscript{4} \\
  \textsuperscript{1} ETH Zurich\\
  \textsuperscript{2} University of Zurich \\
  \textsuperscript{3} University of Potsdam \\
  \textsuperscript{4} Vrije Universiteit Amsterdam \\
  \texttt{\{noraho,fpirovan,ce.zhang\}@inf.ethz.ch},\\\texttt{jaeger@cl.uzh.ch,l.beinborn@vu.nl} }
  
\begin{document}
\maketitle
\begin{abstract}
We analyze if large language models are able to predict patterns of human reading behavior. We compare the performance of language-specific and multilingual pretrained transformer models to predict reading time measures reflecting natural human sentence processing on Dutch, English, German, and Russian texts. This results in accurate models of human reading behavior, which indicates that transformer models implicitly encode relative importance in language in a way that is comparable to human processing mechanisms. We find that BERT and XLM models successfully predict a range of eye tracking features. In a series of experiments, we analyze the cross-domain and cross-language abilities of these models and show how they reflect human sentence processing.
\end{abstract}

\section{Introduction}

When processing language, humans selectively attend longer to the most relevant elements of a sentence \citep{rayner1998eye}. This ability to seamlessly evaluate relative importance is a key factor in human language understanding. It remains an open question how relative importance is encoded in computational language models. Recent analyses conclude that the cognitively motivated ``attention'' mechanism in neural models is not a good indicator for relative importance \cite{jain2019attention}. Alternative methods based on salience \cite{bastings2020elephant}, vector normalization \cite{kobayashi2020attention}, or subset erasure \cite{de2020decisions} are being developed to increase the post-hoc interpretability of model predictions but the cognitive plausibility of the underlying representations remains unclear. 

\begin{figure}[t]
    \centering
    \fbox{\includegraphics[width=0.4\textwidth]{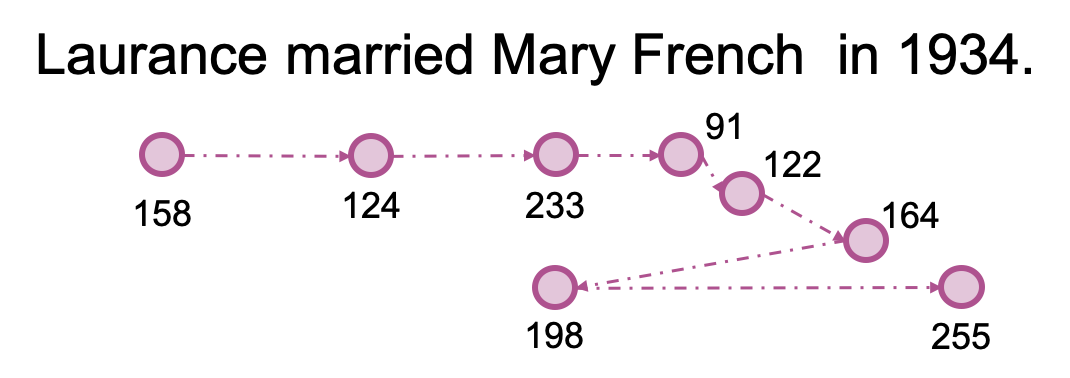}}
    \caption{From the fixation times in milliseconds of a single subject in the ZuCo 1.0 dataset, the feature vector described in Section \ref{sec:et_features} for the wors ``Mary'' would be $[2, 233, 233, 431, 215.5, 1, 1, 1]$.}
    \label{fig:fixations}
\end{figure}

In human language processing, phenomena of relative importance can be approximated indirectly by tracking eye movements and measuring fixation duration \citep{rayner1977visual}. It has been shown that fixation duration and relative importance of text segments are strongly correlated in natural reading, so that direct links can be established on the token level \citep{malmaud2020bridging}. In the example in Figure \ref{fig:fixations}, the newly introduced entity \emph{Mary French} is fixated twice and for a longer duration because it is relatively more important for the reader than the entity \emph{Laurence}, which had been introduced in the previous sentence. Being able to reliably predict eye movement patterns from the language input would bring us one step closer to understand the cognitive plausibility of these models.

Contextualized neural language models are less interpretable than conceptually motivated psycholinguistic models but they achieve high performance in many language understanding tasks and can be fitted successfully to cognitive features such as self-paced reading times and N400 strength \citep{merkx2020comparing}. Moreover, approaches to directly predict cognitive signals (e.g., brain activity) indicate that neural representations implicitly encode similar information as humans \citep{wehbe2014aligning,abnar2019blackbox,sood2020improving,schrimpf2020artificial}. However, it has not been analyzed to which extent transformer language models are able to directly predict human behavioral metrics such as gaze patterns. 
 
The performance of computational models can be improved even further if their inductive bias is adjusted using human cognitive signals such as eye tracking, fMRI, or EEG data \citep{hollenstein2019advancing,toneva2019interpreting,takmaz2020generating}. While psycholinguistic work mainly focuses on very specific phenomena of human language processing that are typically tested in experimental settings with constructed stimuli \citep{hale2017models}, we focus on directly generating token-level predictions from natural reading.

We fine-tune transformer models on human eye movement data and analyze their ability to predict human reading behavior focusing on a range of reading features, datasets, and languages. We compare the performance of monolingual and multilingual transformer models. Multilingual models represent multiple languages in a joint space and aim at a more universal language understanding. As eye tracking patterns are consistent across languages for certain phenomena, we hypothesize that multilingual models might provide cognitively more plausible representations and outperform language-specific models in predicting reading measures. We test this hypothesis on 6 datasets of 4 Indo-European languages, namely English, German, Dutch and Russian.\footnote{Code available on GitHub: \url{https://github.com/DS3Lab/multilingual-gaze}}

We find that pretrained transformer models are surprisingly accurate at predicting reading time measures in four Indo-European languages. Multilingual models show an advantage over language-specific models, especially when fine-tuned on smaller amounts of data. Compared to previous psycholinguistic reading models, the accuracy achieved by the transformer models is remarkable. Our results indicate that transformer models implicitly encode relative importance in language in a way that is comparable to human processing mechanisms. As a consequence, it should be possible to adjust the inductive bias of neural models towards more cognitively plausible outputs without having to resort to large-scale cognitive datasets.

\section{Related Work}

% ET in NLP
Using eye movement data to modify the inductive bias of language processing models has resulted in improvements for several NLP tasks (e.g., \citealt{barrett2016weakly,hollenstein2019entity}).
It has also been used as a supervisory signal in multi-task learning scenarios \citep{klerke2016improving,gonzalez2017using} and as a method to fine-tune the attention mechanism \citep{barrett2018sequence}. We use eye tracking data to evaluate how well transformer language models predict human sentence processing. Therefore, in this section, we discuss previous work on probing transformers models as well as on modelling human sentence processing.

\subsection{Probing Transformer Language Models}

Contextualized neural language models have become increasingly popular, but our understanding of these black box algorithms is still rather limited \citep{gilpin2018explaining}. Current intrinsic evaluation methods do not capture the cognitive plausibility of language models \citep{manning2020emergent,gladkova2016intrinsic}. In previous work of interpreting and probing language models, human behavioral data as well as neuroimaging recordings have been leveraged to understand the inner workings of the neural models. For instance, \citet{ettinger2019bert} explores the linguistic capacities of BERT with a set of psycholinguistic diagnostics. \citet{toneva2019interpreting} propose an interpretation approach by learning alignments between the models and brain activity recordings (MEG and fMRI).
\citet{hao2020probabilistic} propose to evaluate language model quality based on the degree to which they exhibit human-like behavior such as predictability measures collected from human subjects. However, their metric does not reveal any details about the commonalities between the model and human sentence processing.
%Their metric is correlated more closely to modeling reading times than perplexity. 

\begin{table*}[t]
\centering
\small
\begin{tabular}{llrrrrrrr}
\toprule
\textbf{Language} &\textbf{Corpus}  & \textbf{Subjs.} & \textbf{Sents.} & \textbf{Sent. length} & \textbf{Tokens} & \textbf{Types}  & \textbf{Word length} & \textbf{Flesch} \\
\midrule
\multirow{3}{*}{English} & Dundee  & 10 & 2,379 & 21.7 (1--87) & 51,497 & 9,488 & 4.9 (1--20) & 53.3\\
 &GECO  & 14 & 5,373 & 10.5 (1--69) & 56,410 & 5,916 & 4.6 (1--33) & 77.4 \\
 &ZuCo  & 30 & 1,053 & 19.5 (1--68)& 20,545 & 5,560 & 5.0 (1--29) & 50.6\\
\midrule
Dutch & GECO & 19 & 5,190 & 11.64 (1--60) & 59,716 & 5,575 & 4.5 (1--22) & 57.5\\
German & PoTeC & 30 & 97 & 19.5 (5--51) & 1,895 & 847 & 6.5 (2--33) & 36.4\\
Russian & RSC & 103 & 144 & 9.4 (5--13) & 1,357 & 993 & 5.7 (1--18) & 64.7  \\
\bottomrule
\end{tabular}
\caption{Descriptive statistics of all eye tracking datasets.\footnotemark \, Sentence length and word length are expressed as the mean with the min-max range in parentheses. The last column shows the Flesch Reading Ease score \citep{flesch1948new} which ranges from 0 to 100 (higher score indicates easier to read). Adaptations of the Flesch score were used for Dutch (nl), German (de) and Russian (ru) (see Appendix \ref{app:readability}).}

\label{tab:data}
\end{table*}

% Multilingual LMs
The benefits of multilingual models are controversial. Transformer models trained exclusively on a specific language often outperform multilingual models trained on various languages simultaneously, even after fine-tuning. This \textit{curse of multilinguality} \citep{conneau2020unsupervised, vulic2020multi} has been shown for Spanish \citep{canete2020spanish}, 
%Persian \citep{farahani2020parsbert},
Finnish \citep{virtanen2019multilingual} %French \citep{martin2019camembert}
and Dutch \citep{vries_bertje_2019}. 
In this paper we investigate whether a similar effect can be observed when leveraging these models to predict human behavioral measures, or whether in that case the multilingual models provide more plausible representations of human reading due to the common eye tracking effects across languages.

\subsection{Modelling Human Sentence Processing}
% Cognitive plausibility of transformer LMs
Previous work of neural modelling of human sentence processing has focused on recurrent neural networks, since their architecture and learning mechanism appears to be cognitively plausible \citep{keller2010cognitively,michaelov2020well}. However, recent work suggests that transformers perform better at modelling certain aspects of the human language understanding process \citep{hawkins2020investigating}. While \citet{merkx2020comparing} and \citet{wilcox2020predictive} show that the psychometric predictive power of transformers outperforms RNNs on eye tracking, self-paced reading times and N400 strength, they do not directly predict cognitive features. \citet{schrimpf2020artificial} show that contextualized monolingual English models accurately predict language processing in the brain. 

\footnotetext{Note that the exact numbers might differ slightly from the original publications due to different preprocessing methods.}

Context effects are known to influence fixations times during reading \citep{morris1994lexical}. The notion of using contextual information to process language during reading has been well-established in psycholinguistics (e.g., \citealt{inhoff1986parafoveal} and \citealt{jian2013context}). However, to the best of our knowledge, we are the first to study to which extent the representations learned by transformer language models entail these human reading patterns.

% Reading models
Compared to neural models of human sentence processing, we predict not only individual metrics but a range of eye tracking features covering the full reading process from early lexical access to late syntactic processing. By contrast, most models of reading focus on predicting skipping probability \citep{reichle1998toward,matthies2013with,hahn2016modeling}. %These models achieve between 55 and 70\% accuracy when predicting fixation probability. 
%\citet{sood2020improving} propose a visual saliency model to compute the attention scores for task-specific NLP models. 
\citet{sood2020improving} propose a text saliency model which predicts fixation durations that are then used to compute the attention scores in a transformer network.
%Moreover, the results by \citet{sood2020interpreting} suggest that different architectures seem to learn rather different neural attention strategies and similarity of neural to human attention does not guarantee best performance in line with findings by \citet{jain2019attention}. 

\begin{table*}[t]
\centering
\small
\begin{tabular}{llll}
\toprule
\textbf{Short Name} & \textbf{Language} & \textbf{Model Checkpoint} & \textbf{Reference} \\ \midrule
\textsc{bert-nl} & Dutch & \textsc{wietsedv/bert-base-dutch-cased} & \newcite{vries_bertje_2019} \\
\textsc{bert-en} & English & \textsc{bert-base-uncased} & \newcite{Wolf2019HuggingFacesTS} \\
\textsc{bert-de} &German & \textsc{bert-base-german-cased} & \newcite{germanBert} \\
\textsc{bert-ru} &Russian & \textsc{DeepPavlov/rubert-base-cased} & \newcite{yu2019adaptation} \\ 
\textsc{bert-multi} & 104 languages & \textsc{bert-base-multilingual-cased} & \newcite{Wolf2019HuggingFacesTS} \\
\midrule
\textsc{xlm-en} & English & \textsc{xlm-mlm-en-2048} & \newcite{lample2019cross} \\
\textsc{xlm-ende} & English + German & \textsc{xlm-mlm-ende-1024} & \newcite{lample2019cross} \\
\textsc{xlm-17} & 17 languages & \textsc{xlm-mlm-17-1280} & \newcite{lample2019cross} \\
\textsc{xlm-100} & 100 languages & \textsc{xlm-mlm-100-1280} & \newcite{lample2019cross} \\ 
\bottomrule
\end{tabular}
\caption{Pretrained transformer language models analyzed in this work.}
\label{tab:lang-models}
\end{table*}

\section{Data}\label{sec:gaze_data}
We predict eye tracking data only from naturalistic reading studies in which the participants read full sentences or longer spans of naturally occurring text in their own speed. The data from these studies exhibit higher ecological validity than studies which rely on artificially constructed sentences and paced presentation \citep{alday2019m}. 

\subsection{Corpora}
To conduct a cross-lingual comparison, we use eye tracking data collected from native speakers of four languages (see Table \ref{tab:data} for details).
\paragraph{English}
The largest number of eye tracking data sources are available for English. We use eye tracking features from three English corpora: (1) The Dundee corpus \cite{kennedy2003dundee} contains 20 newspaper articles from \textit{The Independent}, which were presented to English native readers on a screen five lines at a time. (2) The GECO corpus \cite{cop2017presenting} contains eye tracking data from English monolinguals reading the entire novel \textit{The Mysterious Affair at Styles} by Agatha Christie. 
%We use the data from the monolingual English speakers. 
The text was presented on the screen in paragraphs. (3) The ZuCo corpus \cite{hollenstein2018zuco,hollenstein2020zuco} includes eye tracking data of full sentences from movie reviews and Wikipedia articles.\footnote{We use Tasks 1 and 2 from ZuCo 1.0 and Task 1 from ZuCo 2.0.}

\paragraph{Dutch} The GECO corpus \cite{cop2017presenting} additionally contains eye tracking data from Dutch readers, which were presented with the same novel in their native language. %Differently from the English part of the corpus, each subject only read half of the novel in Dutch.

\paragraph{German}
The Potsdam Textbook Corpus (PoTeC, \citealt{jaeger2021potsdam}) contains 12 short passages of 158 words on average from college-level biology and physics textbooks, which are read by expert and laymen German native speakers. The full passages were presented on multiple lines on the screen.

\paragraph{Russian}
The Russian Sentence Corpus (RSC, \citealt{laurinavichyute2019russian}) contains 144 naturally occurring sentences extracted from the Russian National Corpus.\footnote{\url{https://ruscorpora.ru}} Full sentences were presented on the screen to monolingual Russian-speaking adults one at a time.

\subsection{Eye Tracking Features}\label{sec:et_features}
A fixation is defined as the period of time where the gaze of a reader is maintained on a single location. Fixations are mapped to words by delimiting the boundaries around the region on the screen belonging to each word \emph{w}. A word can be fixated more than once.
For each token \emph{w} in the input text, we predict the following eight eye tracking features that encode the full reading process from early lexical access up to subsequent syntactic integration.

\begin{figure*}[t]
    \centering
    \includegraphics[width=0.43\textwidth]{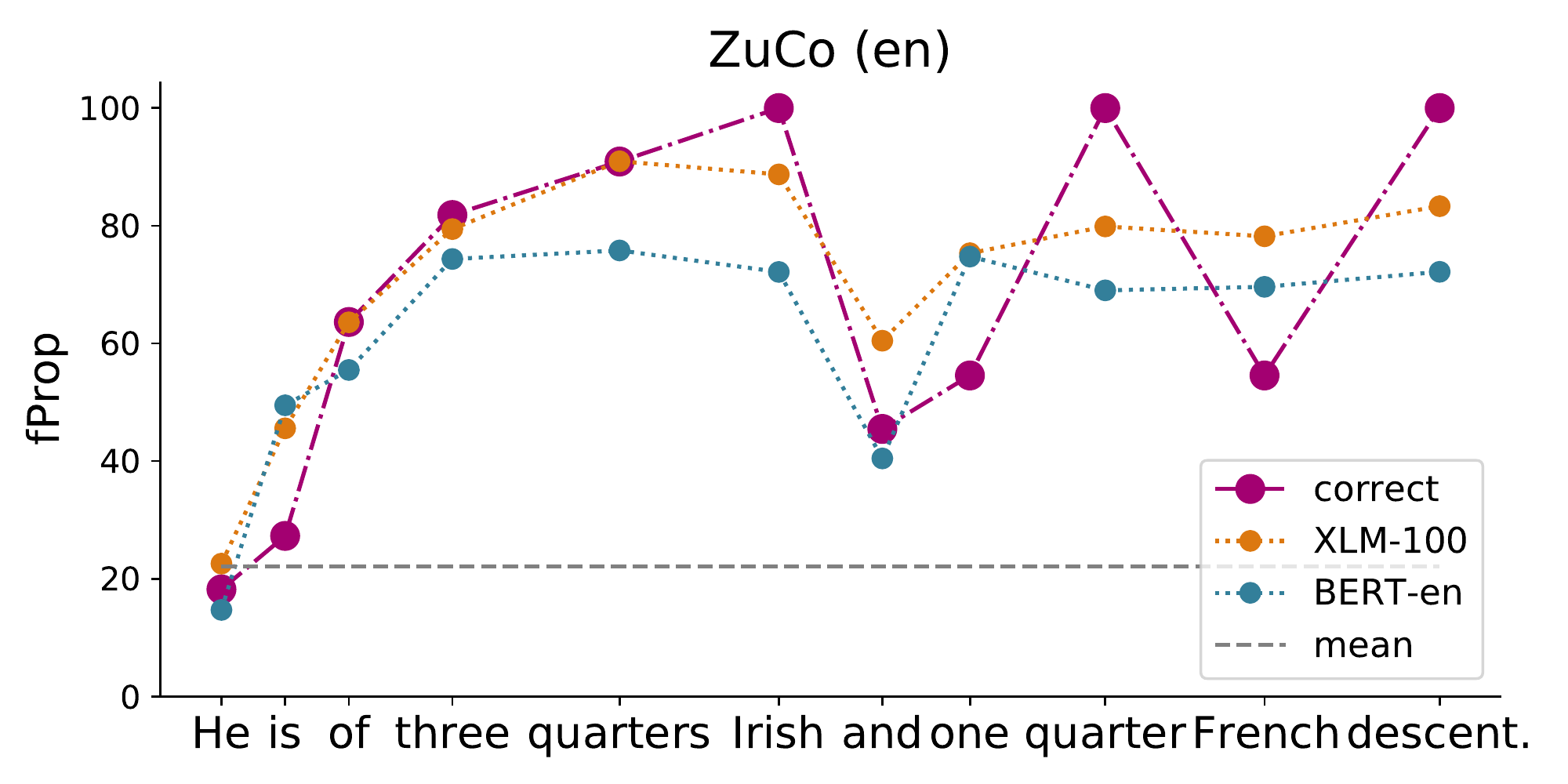}
    \includegraphics[width=0.52\textwidth]{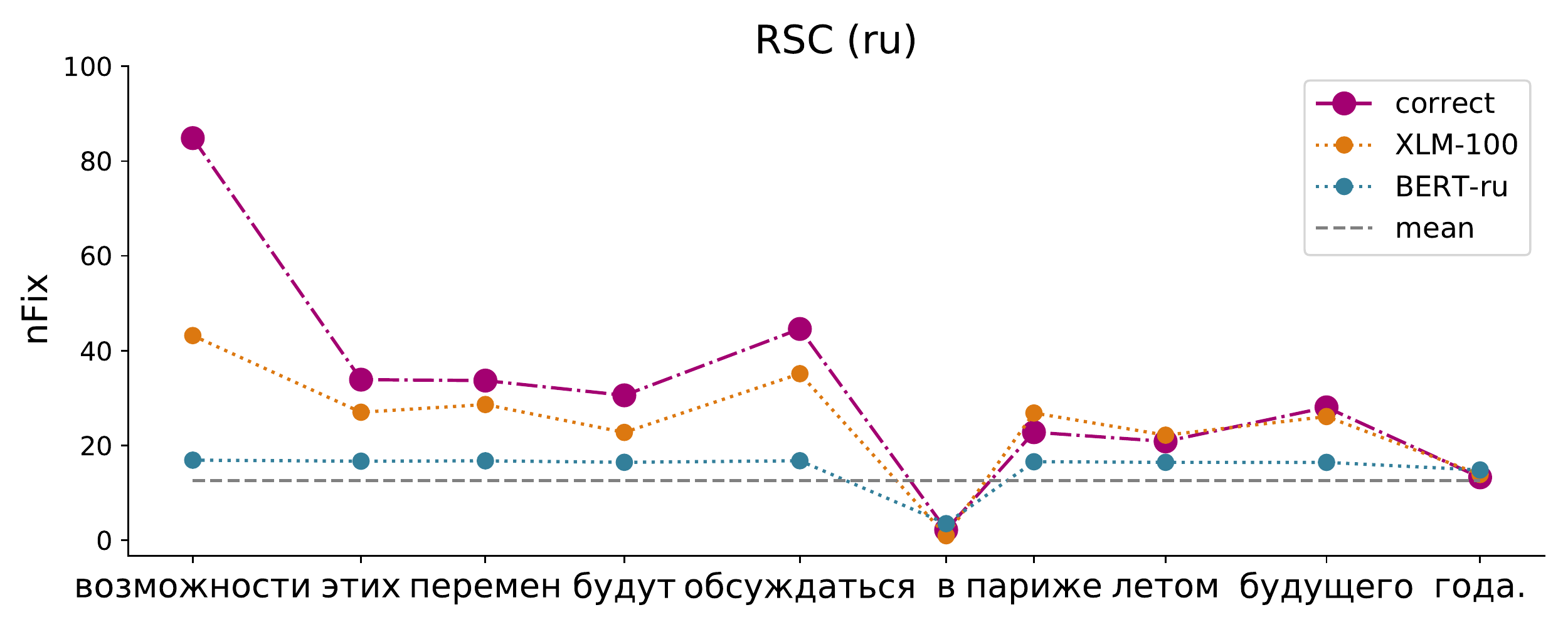}
    \caption{True and predicted feature values for two example sentences. On the left the fixation proportion (\textsc{fProp}) values for an English sentence from the ZuCo dataset, and on the right the number of fixations (\textsc{nFix}) values for a Russian sentence from the RSC dataset.}
    \label{fig:example-predictions}
\end{figure*}

\paragraph{Word-level characteristics} We extract basic features that encode \textit{word-level} characteristics: (1) number of fixations (\textsc{nFix}), the number of times a subject fixates \emph{w}, averaged over all subjects; (2) mean fixation duration (\textsc{MFD}), the average fixation duration of all fixations made on \emph{w}, averaged over all subjects;
(3) fixation proportion (\textsc{fProp}), the number of subjects that fixated \emph{w}, divided by the total number of subjects.

\paragraph{Early processing} We also include features to capture the \textit{early} lexical and syntactic processing, based on the first time a word is fixated: (4) first fixation duration (\textsc{FFD}), the duration, in milliseconds, of the first fixation on \emph{w}, averaged over all subjects; (5) first pass duration (\textsc{FPD}), the sum of all fixations on \emph{w} from the first time a subject fixates \emph{w} to the first time the subject fixates another token, averaged over all subjects.

\paragraph{Late processing} Finally, we also use measures reflecting the \textit{late} syntactic processing and general disambiguation, based on words which were fixated more than once: (6) total reading time (\textsc{TRT}), the sum of the duration of all fixations made on \emph{w}, averaged over all subjects; 
(7) number of re-fixations (\textsc{nRefix}), the number of times \emph{w} is fixated after the first fixation, i.e., the maximum between 0 and the \textsc{nFix}-1, averaged over all subjects; (8) re-read proportion (\textsc{ReProp}), the number of subjects that fixated \emph{w} more than once, divided by the total number of subjects.

The values of these eye tracking features vary over different ranges (see Appendix \ref{app:et-data}).
\textsc{FFD}, for example, is measured in milliseconds, and average values are around 200 ms, whereas \textsc{ReProp} is a proportional measure, and therefore assumes floating-point values between 0 and 1. We standardize all eye tracking features independently (range: 0--100), so that the loss can be calculated uniformly over all feature dimensions. 

Eye movements depend on the stimulus and are therefore language-specific but there exist universal tendencies which remain stable across languages \citep{liversedge2016universality}. For example, the average fixation duration in reading ranges from 220 to 250 ms independent of the language. Furthermore, word characteristics such as word length, frequency and predictability affect fixation duration similarly across languages but the effect size depends on the language and the script \citep{laurinavichyute2019russian,bai2008reading}.  
The word length effect, i.e., the fact that longer words are more likely to be fixated, can be observed across all four languages included in this work (see Appendix \ref{app:et-data}). 

\section{Language Models}
We compare the ability to predict eye tracking features in two models: BERT and XLM. Both models are trained on the transformer architecture \citep{vaswani2017attention} and yield state-of-the-art results for a wide range of NLP tasks \citep{liang2020xglue}. The multilingual BERT model simply concatenates the Wikipedia input from 104 languages and is optimized by performing masked token and next sentence prediction as in the monolingual model \citep{devlin2019bert} without any cross-lingual constraints. In contrast, XLM adds a translation language modeling objective, by explicitly using parallel sentences in multiple languages as input to facilitate cross-lingual transfer \citep{lample2019cross}. Both BERT and XLM use subword tokenization methods to build shared vocabulary spaces across languages.

We use the pretrained checkpoints from the HuggingFace repository for monolingual and multilingual models (details in Table \ref{tab:lang-models}).\footnote{\url{https://huggingface.co/transformers/pretrained_models.html}}

\begin{table*}[t]
\centering
\small
\begin{tabular}{lcccc}
\toprule
\textbf{Model} & \textbf{Dundee (en)} & \textbf{GECO (en)} & \textbf{ZuCo (en)} & \textbf{ALL (en)} \\\midrule
\textsc{bert-en} & 92.63 (0.05) & 93.68 (0.14) & 93.42 (0.02) & 93.71 (0.06) \\
%\textsc{bert-en-cased} & 92.73 (0.07) & 93.71 (0.13) & 93.83 (0.09) & 93.73 (0.06) \\
\textsc{bert-multi} & 92.73 (0.06) & 93.73 (0.12) & 93.74 (0.05) & 93.74 (0.07) \\\hline
\textsc{xlm-en} & 90.41 (2.16) & 91.15 (1.42) & 92.03 (2.11) & 90.88 (1.50) \\
\textsc{xlm-ende} & 92.79 (0.15) & \textbf{93.89} (0.12) & 93.76 (0.15) & \textbf{93.96} (0.08) \\
\textsc{xlm-17} & 92.11 (1.68) & 91.79 (1.75) & 92.05 (2.25) & 93.80 (0.38) \\
\textsc{xlm-100} & \textbf{92.99} (0.05) & 93.04 (1.40) & \textbf{93.97} (0.09) & \textbf{93.96} (0.06) \\\bottomrule
\end{tabular}
\caption{Prediction accuracy over all eye tracking features for the English corpora, including the concatenated dataset. Standard deviation is reported in parentheses.}
\label{tab:res-overall-en}
\end{table*}

\begin{table*}[t]
\centering
\small
\begin{tabular}{lcccc}
\toprule
\textbf{Model} & \textbf{GECO (nl)} & \textbf{PoTeC (de)} & \textbf{RSC (ru)} & \textbf{ALL-LANGS} \\\midrule
\textsc{bert-nl}& 91.81 (0.23) &--  &-- &--  \\
\textsc{bert-de} &--  & 78.38 (1.69) &--  &--  \\
\textsc{bert-ru} &--  &--  & 78.73 (1.38) &--  \\
\textsc{bert-multi} & 91.90 (0.16) & 76.86 (2.42) & 76.54 (3.59) & 94.72 (0.07) \\\midrule
\textsc{xlm-ende} &--  & 80.94 (0.88) &--  &--  \\
\textsc{xlm-17} & 91.04 (0.70) & 86.26 (1.31) & 90.96 (3.96) &  94.46 (0.83)\\
\textsc{xlm-100} & \textbf{92.31} (0.22) & \textbf{86.57} (0.54) & \textbf{94.70} (0.60) & \textbf{94.94} (0.11)\\\bottomrule
\end{tabular}
\caption{Prediction accuracy over all eye tracking features for the Dutch, German and Russian corpora, and for all four languages combined in a single dataset. Standard deviation is reported in parentheses.}
\label{tab:res-overall-langs}
\end{table*}

\section{Method}

We fine-tune the models described above on the features extracted from the eye tracking datasets. The eye tracking prediction uses a model for token regression, i.e., the pretrained language models with a linear dense layer on top of it. The final dense layer is the same for all tokens, and performs a projection from the dimension of the hidden size of the model (e.g., 768 for \textsc{bert-en} or 1,280 for \textsc{xlm-100}) to the dimension of the eye tracking feature space (8, in our case). The model is trained for the regression task using the \textit{mean squared error} (MSE) loss.

\paragraph{Training Details}
We split the data into 90\% training data, 5\% validation and 5\% test data. We initially tuned the hyper-parameters manually and set the following values for all models: We use an AdamW optimizer \cite{loshchilov2018decoupled} with a learning rate of $0.00005$ and a weight decay of $0.01$. The batch size varies depending on the model dimensions (see Appendix \ref{app:batch}).
We employ a linear learning rate decay schedule over the total number of training steps. We clip all gradients exceeding the maximal value of 1. We train the models for 100 epochs, with early stopping after 7 epochs without an improvement on the validation accuracy.

\paragraph{Evaluation Procedure}
As the features have been standardized to the range 0--100, the \textit{mean absolute error} (MAE) can be interpreted as a percentage error. For readability, we report the \textit{prediction accuracy} as $100-$MAE in all experiments. The results are averaged over batches and over 5 runs with varying random seeds. For a single batch of sentences, the overall MAE is calculated by concatenating the words in each sentence and the feature dimensions for each word, and padding to the maximum sentence length. The per-feature MAE is calculated by concatenating the words in each sentence.
For example, for a batch of \textit{B} sentences, each composed of \textit{L} words, and \textit{G} eye tracking features per word, the overall MAE is calculated over a vector of \textit{B*L*G} dimensions. In contrast, the MAE for each individual feature is calculated over a vector of \textit{B*L} dimensions.

\begin{figure*}[t]
    \centering
    \includegraphics[width=0.39\textwidth]{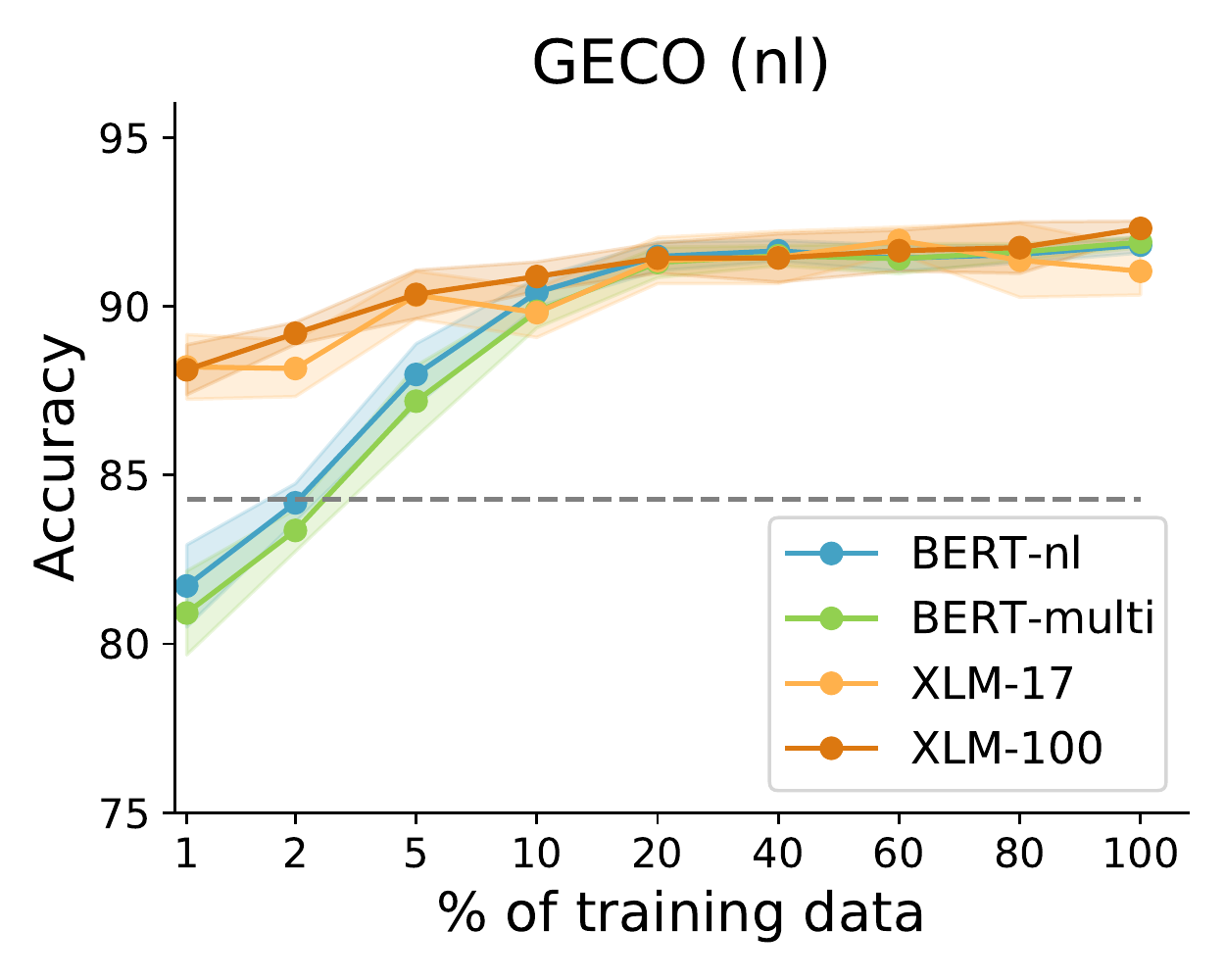}
    \hspace{0.1\textwidth}
    \includegraphics[width=0.39\textwidth]{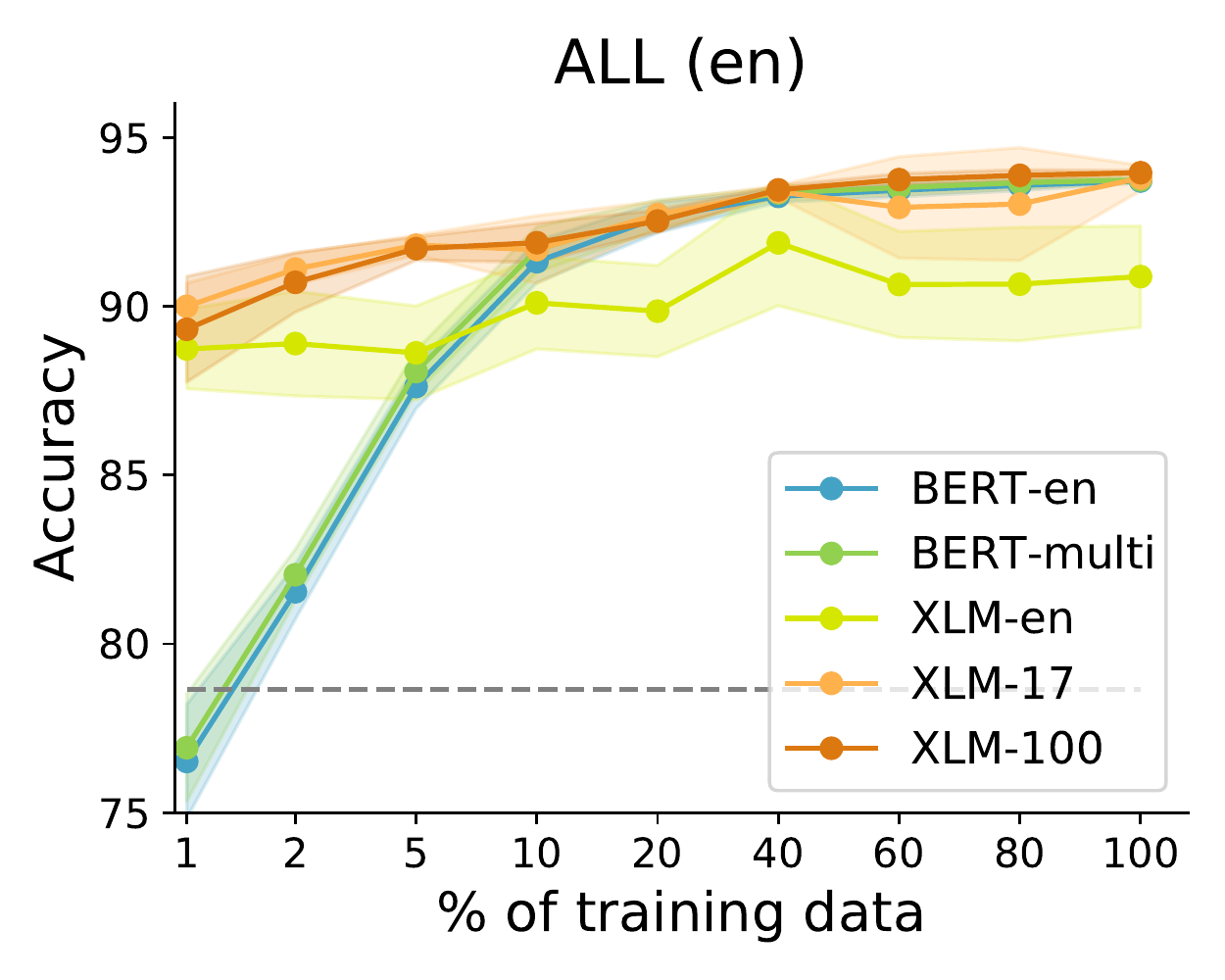}
    \caption{Data ablation study for Dutch and English. The results are aggregated over all eye tracking features. In addition to the mean across five runs, the shaded areas represent the standard deviation. The dashed line is the result of the pretrained \textsc{bert-multi} model without fine-tuning. Results are aggregated over all eye tracking features.
    }
    \label{fig:data-ablation}
\end{figure*}

\section{Results \& Discussion}

Tables \ref{tab:res-overall-en} and \ref{tab:res-overall-langs} show that all models predict the eye tracking features with more than 90\% accuracy for English and Dutch. 
For English, the BERT models yield high performance on all three datasets with standard deviations below 0.15.
The results for the XLM models are slightly better on average but exhibit much higher standard deviations. 
Similar to the results presented by \newcite{lample2019cross}, we find that more training data from \textit{multiple} languages improves prediction performance. For instance, the \textsc{xlm-100} model achieves higher accuracy than the \textsc{xlm-17} model in all cases. For the smaller non-English datasets, PoTeC (de) and RSC (ru), the multilingual XLM models clearly outperform the monolingual models. For the English datasets, the differences are minor. 

\begin{figure}[t]
    \centering
    \includegraphics[width=0.49\textwidth,trim=10 0 0 20]{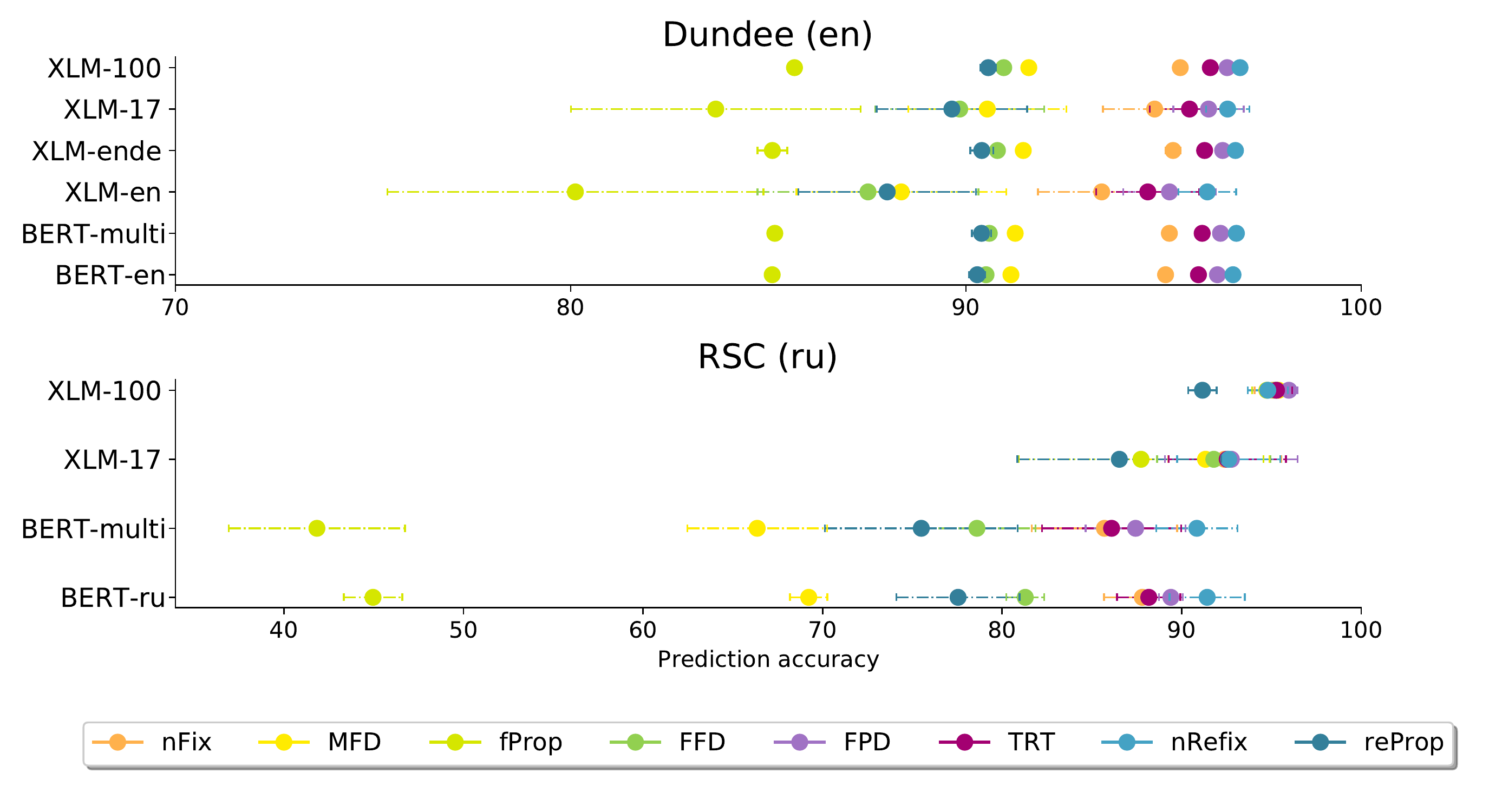}
    \caption{Results of individual eye tracking features for all models on the Dundee and RSC corpora.}
    \label{fig:results-features-all-models}
\end{figure}

\begin{figure}[t]
    \centering
    \includegraphics[width=0.49\textwidth,trim=0 0 0 20]{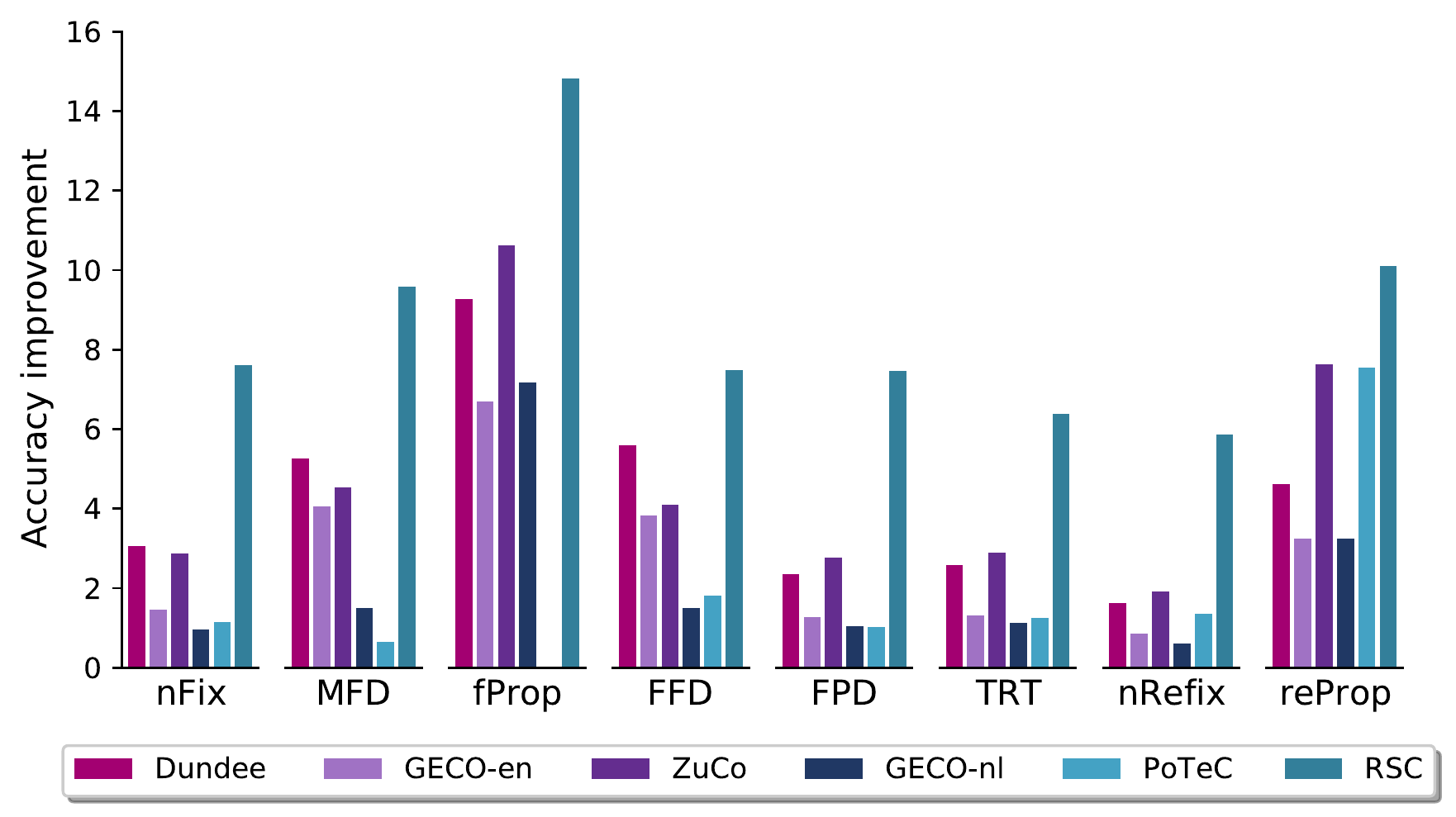}
    \caption{Improvement of prediction accuracy for the \textsc{xlm-100} model relative to the mean baseline for each eye tracking feature.}
    \label{fig:results-features-langs-selected}
\end{figure}

\begin{figure*}[ht]
    \centering
    \includegraphics[trim= 15 15 15 15,width=0.9\textwidth]{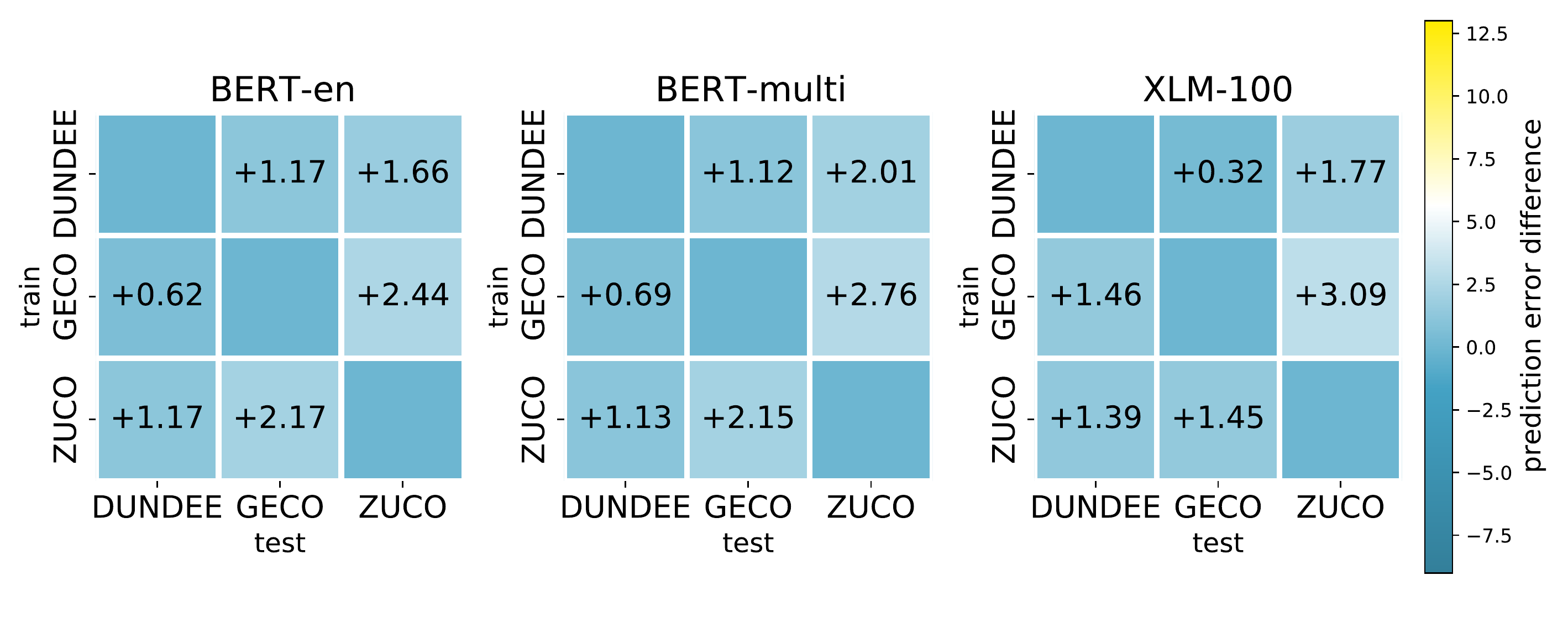}
    \caption{Cross-domain evaluation on pretrained English models. The results are expressed as the difference in the prediction error compared to the in-domain prediction. A smaller error (i.e., a color more similar to the color of the diagonal) represents better domain adaptation.}
    \label{fig:cross-domain}
\end{figure*}

\begin{figure*}[ht]
    \centering
    \includegraphics[trim= 15 15 15 15, width=0.9\textwidth]{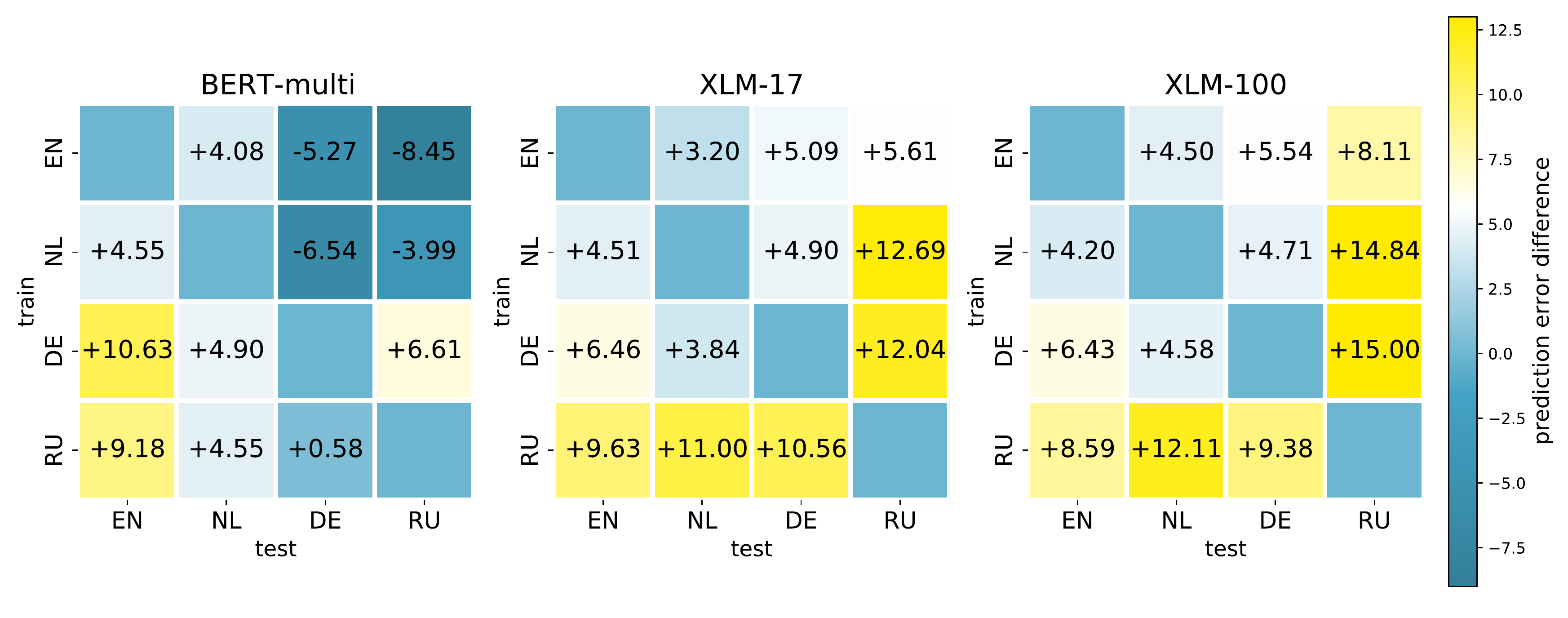}
    \caption{Cross-language evaluation on multilingual models across English, Dutch, German and Russian data. The results are expressed as the difference in the prediction error compared to the prediction on the same language. A smaller error (i.e., a color more similar to the color of the diagonal) represents better language transfer.}
    \label{fig:cross-lang}
\end{figure*}

\paragraph{Size Effects}
More training data results in higher prediction accuracy even when the eye tracking data comes from various languages and was recorded in different reading studies by different devices (ALL-LANGS, fine-tuning on the data of all four languages together). However, merely adding more data from the same language (ALL (en), fine-tuning on the English data from Dundee, GECO and ZuCo together) does not result in higher performance. 

To analyze this further, we perform an ablation study on varying amounts of training data. The results are shown in Figure \ref{fig:data-ablation} for Dutch and English. The performance of the XLM models remains stable even with a very small percentage of eye tracking data. The performance of the BERT models, however, drops drastically when fine-tuning on less than 20\% of the data. Similar to \citet{merkx2020comparing} and \citet{hao2020probabilistic} we find that the model architecture, along
with the composition and size of the training corpus have a significant impact on the psycholinguistic modeling
performance.

\paragraph{Eye Tracking Features}
The accuracy results are averaged over all eye tracking features. For a better understanding of the prediction output, we plot the true and the predicted values of two selected features (\textsc{fProp} and \textsc{nFix}) for two example sentence in Figure \ref{fig:example-predictions}. In both examples, the model predictions strongly correlate with the true values. The difference to the mean baseline is more pronounced for the \textsc{fixProp}feature. 

Figure \ref{fig:results-features-all-models} presents the quantitative differences across models in predicting the individual eye tracking features.\footnote{Plots for the remaining datasets are in Appendix \ref{app:feat-results}} Across all datasets, first pass duration (\textsc{FPD}) and number of re-fixations (\textsc{nRefix}) are the most accurately predicted features. Proportions (\textsc{fProp} and \textsc{reProp}) are harder to predict because these features are even more dependent on subject-specific characteristics. Nevertheless, when comparing the prediction accuracy of each eye tracking feature to a baseline which always predicts the mean values, the predicted features \textsc{fProp} and \textsc{reProp} achieve the largest improvements relative to the mean baseline. See Figure \ref{fig:results-features-langs-selected} for a comparison between all features for the best performing model \textsc{xlm-100} on all six datasets. 

\paragraph{Performance of Pretrained Models} To test the language models' abilities on predicting human reading behavior only from pretraining on textual input, we take the provided model checkpoints and use them to predict the eye tracking features without any fine-tuning.
The detailed results are presented in Appendix \ref{app:pretrained}. The achieved accuracy aggregated over all eye tracking features lies between 75-78\% for English. For Dutch, the models achieve 84\% accuracy but for Russian merely 65\%. Across the same languages the results between the different language models are only minimal. However, on the individual eye tracking features, the pretrained models do not achieve any improvements over the mean baseline (see Appendix \ref{app:pretrained}).

\section{Data Sensitivity}
For the main experiment, we always tested the models on held-out data from the same dataset. In this section, we examine the influence of dataset properties (text domain and language) on the prediction accuracy. In a second step, we analyze the influence of more universal input characteristics (word length, text readability). 

\subsection{Cross-Domain Evaluation}
 Figure \ref{fig:cross-domain} shows the results when evaluating the eye tracking predictions on out-of-domain text for the English datasets. For instance, we fine-tune the model on the newspaper articles of the Dundee corpus and test on the literary novel of the GECO corpus. We can see that the overall prediction accuracy across all eye tracking features is constantly above. 90\% in all combinations. This shows that our eye tracking prediction model is able to generalize across domains. We find that the cross-domain capabilities of BERT are slightly better than for XLM. \textsc{bert-en} performs best in the cross-domain evaluation, possibly because its training data is more domain-general since it includes text from Wikipedia and books.

\subsection{Cross-Language Evaluation}
Figure \ref{fig:cross-lang} shows the results for cross-language evaluation to probe the language transfer capabilities of the multilingual models. We test models fine-tuned on language A on the test set of language B. It can be seen that \textsc{bert-multi} generalizes better across languages than the XLM models. 
This might be due to the fact that the multilingual BERT model is trained on one large vocabulary of many languages but the XLM models are trained with a cross-lingual objective and language information. Hence, during fine-tuning on eye tracking data from one language the XLM models lose some of their cross-lingual abilities.
Our results are in line with \citet{pires2019multilingual} and \citet{karthikeyan2020cross}, who showed that BERT learns multilingual representations in more than just a shared vocabulary space but also across scripts. When fine-tuning \textsc{bert-multi} on English or Dutch data and testing on Russian, we see surprisingly high accuracy across scripts, even outperforming the in-language results. The XLM models, however, show the expected behavior where transferring within the same script (Dutch, English, German) works much better than transferring between the Latin and Cyrillic script (Russian). 

\subsection{Input Characteristics}

\begin{figure}[t]
    \centering
    \includegraphics[width=0.45\textwidth]{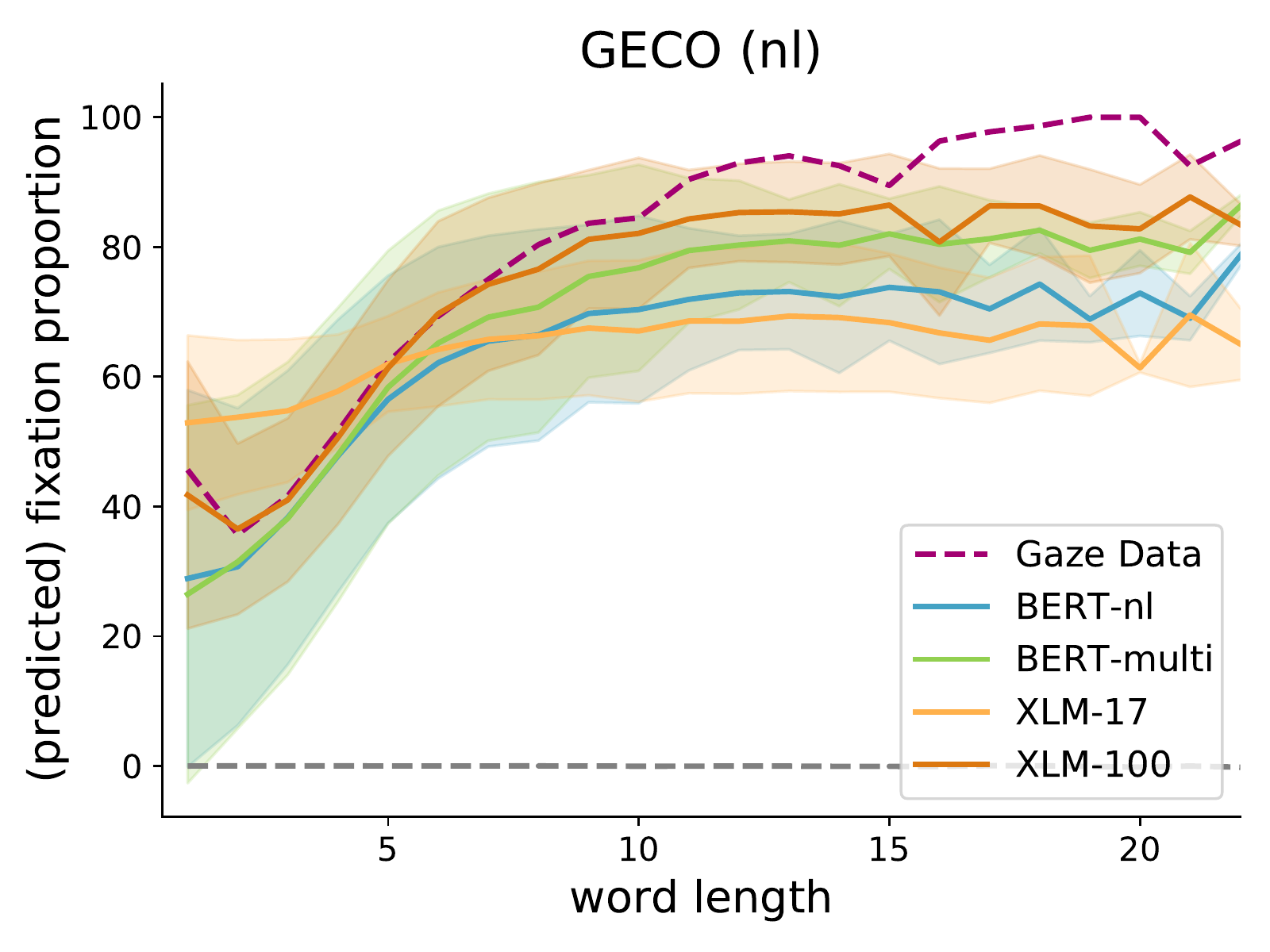}
    \caption{Prediction accuracy of \textsc{fProp} with respect to word length. The gray dashed line is the result of the pretrained \textsc{bert-multi} model without fine-tuning.}
    \label{fig:fix-prob-pred}
\end{figure}

Gaze patterns are strongly correlated with word length. 
Figure \ref{fig:fix-prob-pred} shows that the models accurately learn to predict higher fixation proportions for longer words. We observe that the predictions of the \textsc{xlm-100} model follow the trend in the original data most accurately. Similar patterns emerge for the other languages (see Appendix \ref{app:word-len}). Notably, the pretrained models before fine-tuning do not reflect the word length effect.

\begin{figure}[t]
    \centering
    \includegraphics[width=0.47\textwidth]{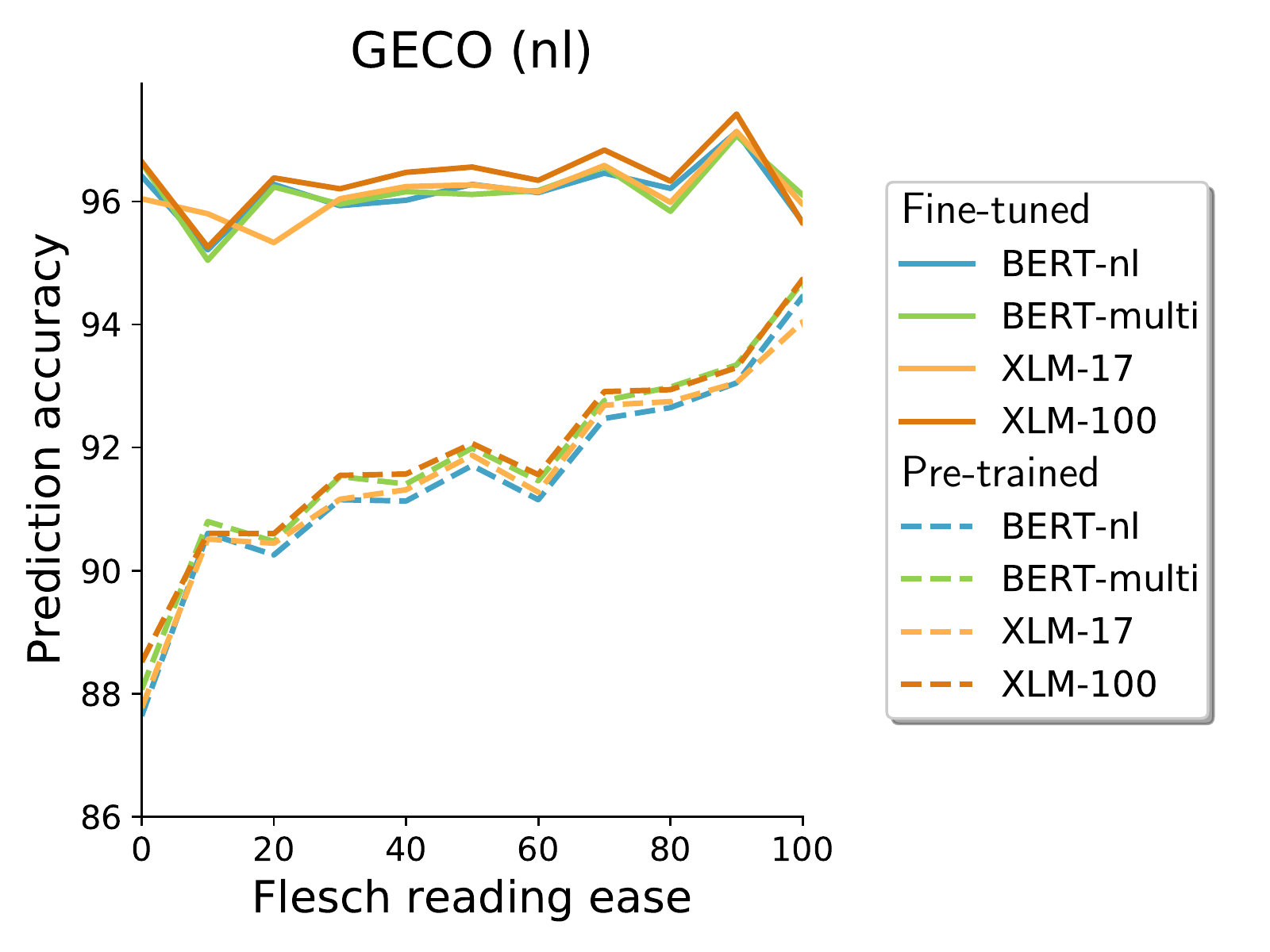}
    \includegraphics[width=0.47\textwidth]{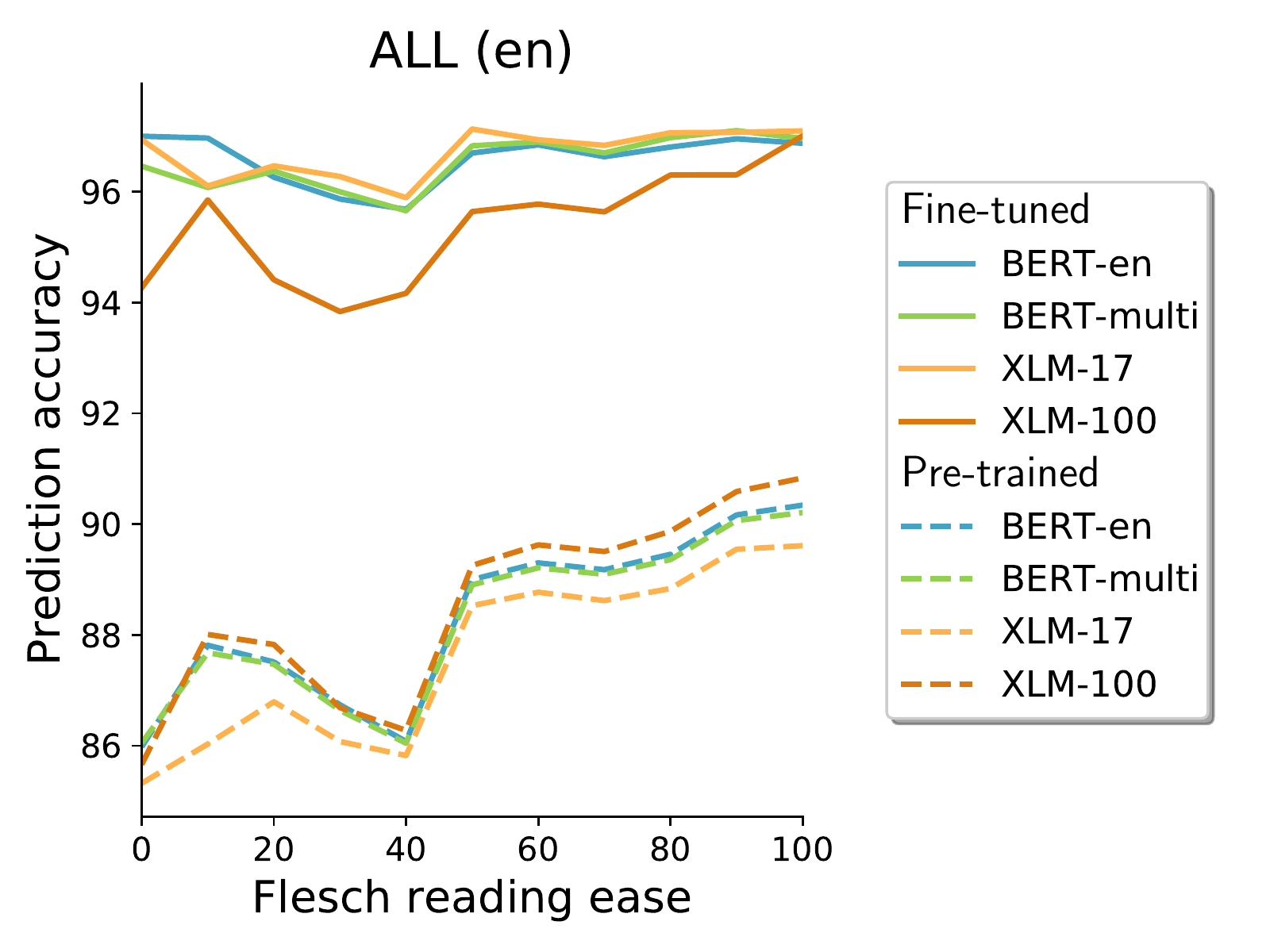}
    \caption{Prediction accuracy for \textsc{nFix} relative to the Flesch reading ease score of the sentence. A higher Flesch score indicates that a sentence is easier to read. The dashed lines show the results of the pretrained language models without fine-tuning on eye tracking data.}
    \label{fig:flesch-analysis}
\end{figure}

On the sentence level, we hypothesize that eye tracking features are easier to predict for sentences with a higher readability. 
Figure \ref{fig:flesch-analysis} shows the accuracy for predicting the number of fixations (\textsc{nFix}) in a sentence relative to the Flesch reading ease score. Interestingly, the pretrained models without fine-tuning conform to the expected behavior and show a consistent increase in accuracy for sentences with a higher reading ease score. After fine-tuning on eye tracking data, this behavior is not as visible anymore since the language models achieve constantly high accuracy independent of the readability of the sentences.

These results might be explained by the nature of the Flesch readability score, which is based only on the structural complexity of the text (see Appendix \ref{app:readability} for a description of the Flesch Reading Ease score). Our results indicate that language models trained purely on textual input are more calibrated towards such structural characteristics, i.e., the number of syllables in a word and the number of words in a sentences. Hence, the Flesch reading ease score might not be a good approximation for text readability. In future work, comparing eye movement patterns and text difficulty should rely on readability measures that take into account lexical, semantic, syntactic, and discourse features. This might reveal deviating patterns between pretrained and fine-tuned models.

Our analyses indicate that the models learn to take properties of the input into account when predicting eye tracking patterns. These processing strategies are similar to those observed in humans. Nevertheless, the connection between readability and relative importance in text needs to be analysed in more detail to establish how well these properties are learned by the language models.

\section{Conclusion}

While the superior performance of pretrained transformer language models has been established, we have yet to understand to which extent these models are comparable to human language processing behavior. We take a step in this direction by fine-tuning language models on eye tracking data to predict human reading behavior. 

We find that both monolingual and multilingual models achieve surprisingly high accuracy in predicting a range of eye tracking features across four languages. Compared to the XLM models, \textsc{bert-multi} is more robust in its ability to generalize across languages, without being explicitly trained for it. In contrast, the XLM models perform better when fine-tuned on less eye tracking data. Generally, fixation duration features are predicted more accurately than fixation proportion, possibly because the latter show higher variance across subjects. We observe that the models learn to reflect characteristics of human reading such as the word length effect and higher accuracy in more easily readable sentences.

%The ability of transformer models to achieve such high results in modelling human sentence processing indicates that we can learn more about the cognitive plausibility of these models by predicting behavioral metrics.

%The ability of transformer models to achieve such high results in modelling reading behavior indicates that we can learn more about the commonalities between language models and human sentence processing by predicting behavioral metrics.

The ability of transformer models to achieve such high results in modelling reading behavior indicates that we can learn more about the commonalities between language models and human sentence processing. By predicting behavioral metrics such as eye tracking features we can investigate the cognitive plausibility within these models to adjust or intensify the human inductive biases.
%need to learn more about the cognitive plausibility within these models to adjust or intensify the human inductive biases.

\clearpage

\section*{Acknowledgements}

Lena Jäger was partially funded by the German Federal Ministry of Education and Research under grant 01|S20043. 

\bibliography{anthology,eeg-eyemove-nlp-oct2017}
\bibliographystyle{acl_natbib}

\clearpage

\appendix

%\section{Example Appendix}
%\label{sec:appendix}

\section{Eye Tracking Data}\label{app:et-data}

Table \ref{tab:et-features} presents information about the range of the eye tracking features.

\begin{figure}[h]
    \centering
    \includegraphics[width=0.49\textwidth]{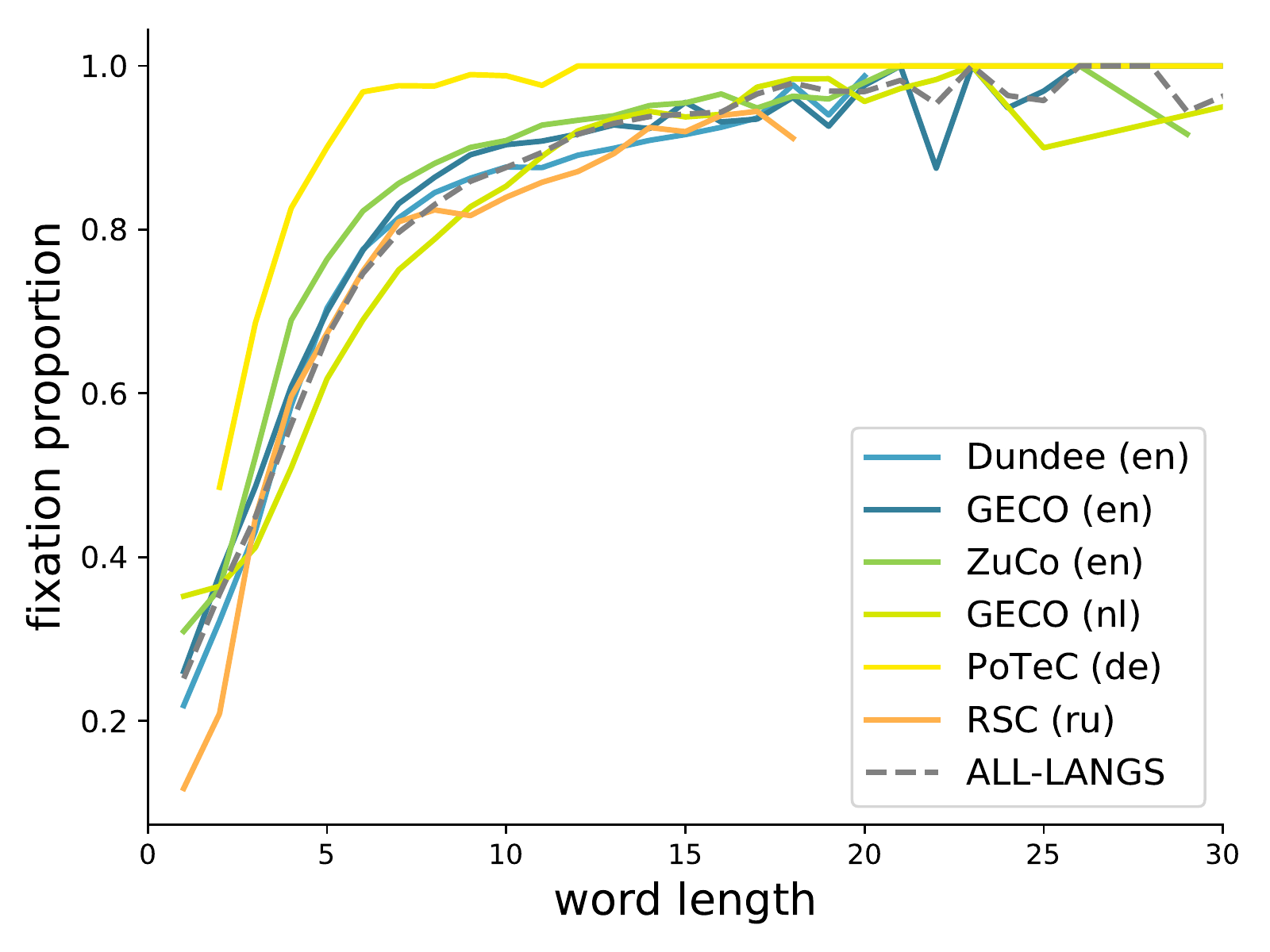}
    \caption{Word length effect on all datasets in all four languages.}
    \label{fig:word-length}
\end{figure}

\noindent Figure \ref{fig:word-length} shows the word length effect found in eye tracking data recorded during reading. i.e., the fact that longer words are more likely to be fixated. This effect is observable across all languages. 

\begin{figure*}[h]
    \centering
    \includegraphics[width=0.89\textwidth]{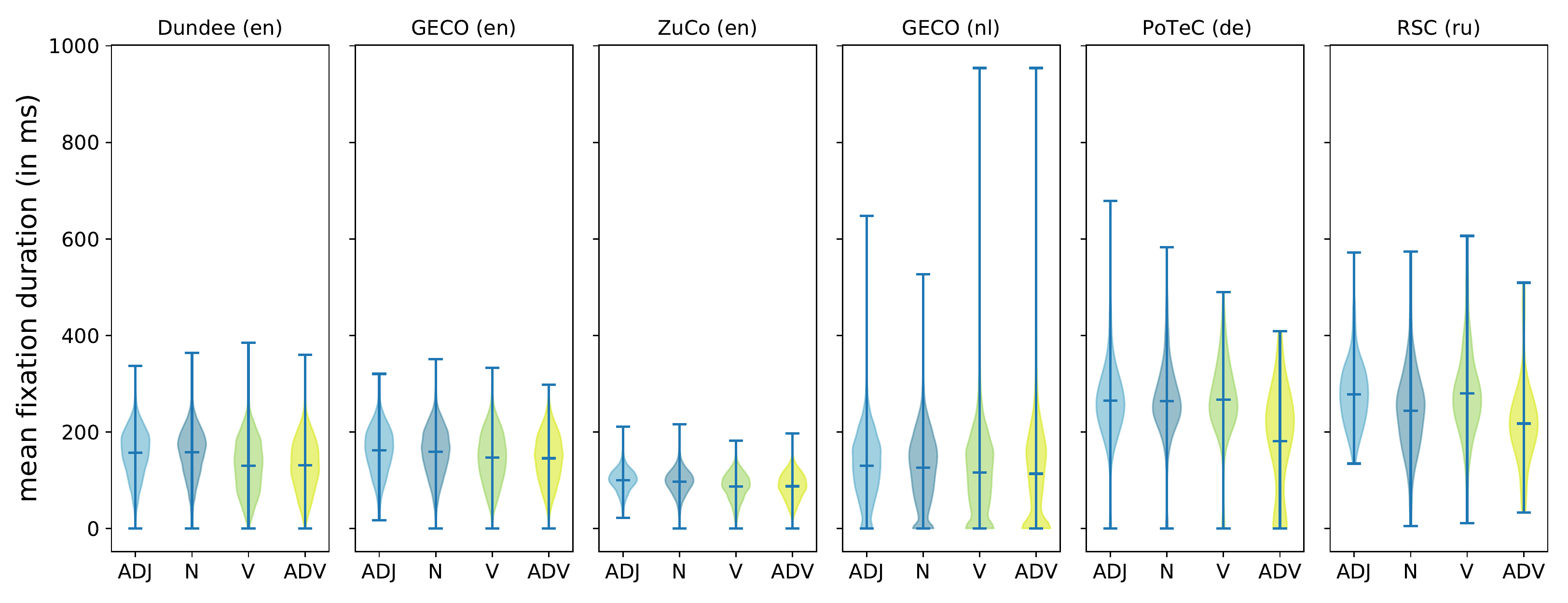}
    \caption{Mean fixation duration (\textsc{MFD}) for the most common parts of speech across all six datasets.}
    \label{fig:pos}
\end{figure*}

\noindent Figure \ref{fig:pos} shows the mean fixation duration (\textsc{MFD}) for adjectives, nouns, verbs, and adverbs for all six datasets. We use spacy\footnote{\url{spaCy.io}} to perform part-of-speech tagging for our analyses. For Russian we load an externally trained model\footnote{\url{https://github.com/buriy/spacy-ru}}, for Dutch, English and German we use the provided pretrained models. 

\begin{figure*}[h]
    \centering
    \includegraphics[width=0.49\textwidth]{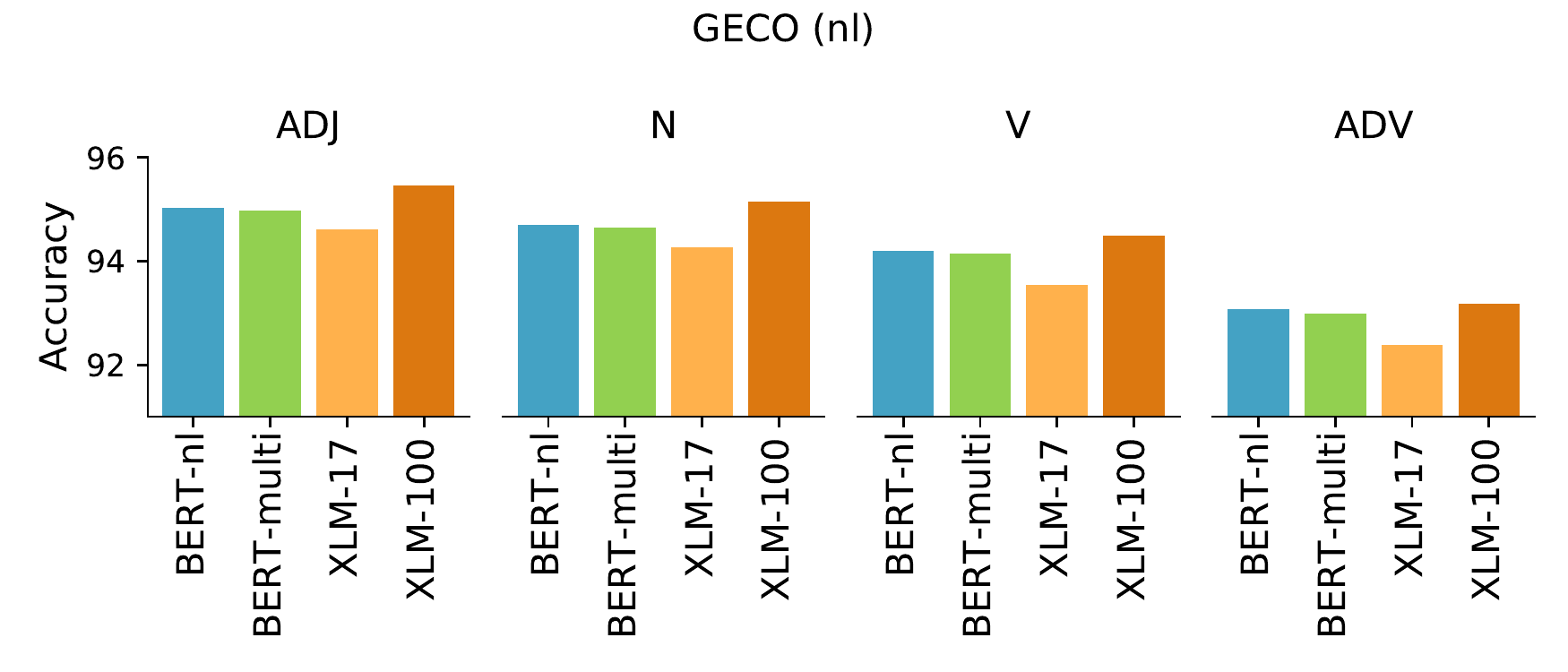}
    \includegraphics[width=0.49\textwidth]{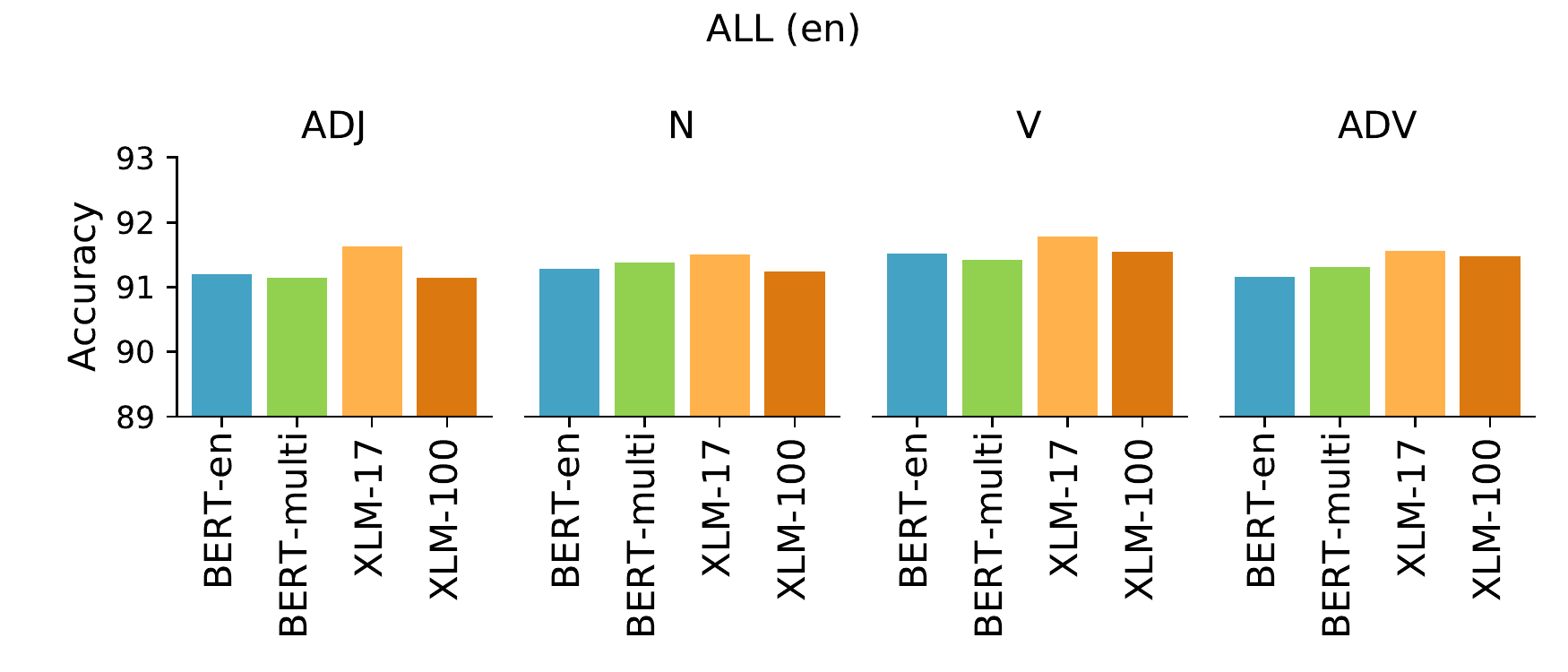}
    \caption{Accuracy of the language models predicting the mean fixation duration (\textsc{MFD}) across various parts of speech for Dutch (left) and English (right).}
    %English = ALL-EN
    \label{fig:pos-acc}
\end{figure*}

\noindent Figure \ref{fig:pos-acc} shows an additional analysis where we explore which parts-of-speech can be predicted more accurately by the language models.

\section{Readability Scores}\label{app:readability}

We use the Flesch Reading Easy score \citep{flesch1948new} to define the readability of the English text in the eye tracking corpora. This score indicates how difficult a text passage is to understand. Since this score relies on language-specific weighting factors, we apply the Flesch Douma adaptation for Dutch \citep{douma1960leesbaarheid}, the adaptation by \citet{amstad1978verstandlich} for German, and the adaptation by \citet{oborneva2006automatic} for Russian.

\section{Implementation Details}

\subsection{Tokenization} 
When using BERT or XLM for token classification or regression, a pressing implementation issue is represented by the subword tokenizers employed by the models.
This tokenizer, in fact, handles unknown tokens by recursively splitting every word until all subtokens belong to its vocabulary.
For example, the name of the Greek mythological hero ``Philammon'' is tokenized into the three subtokens ``[`phil', `\#\#am', `\#\#mon']''.
In this case, our models for token regression would produce an eight-dimensional output for all three subtokens, and we had the choice as to what to do in order to compute the loss, having only one target for the full word ``Philammon''.
We chose to compute the loss only with respect to the first subtoken.

\subsection{Training Setup}\label{app:batch}

As described in the main paper, all experiments are run over 5 random seeds, which are $\{12,79,237,549,886\}$.

\noindent All models were fine-tuned on a single GPU Titan X with 12 GB memory. Due to memory restrictions of the GPUs and the dimensions of the language models, the batch size was adapted as needed. Table \ref{tab:btach-size} shows the batch sizes for each model.

\begin{table}[h]
\centering
\begin{tabular}{lc}
\toprule
\textbf{Model} & \textbf{Batch size} \\\midrule
\textsc{bert-en}, \textsc{bert-nl}, & 16 \\
\textsc{bert-multi} &  \\\hline
\textsc{bert-de}, \textsc{bert-ru}, & 8\\ 
\textsc{xlm-ende}, \textsc{xlm-17},&  \\ 
\textsc{xlm-100} &  \\\hline
\textsc{xlm-en} & 2 \\
\bottomrule
\end{tabular}
\caption{Batch sizes used for each of the language models.}
\label{tab:btach-size}
\end{table}

\noindent On average the validation accuracy of BERT models stops improving after $\sim50$ epochs, while the XLM models only take $\sim10$ epochs. There is no noteworthy difference in training speed between monolingual and multilingual models.

\section{Detailed Results}

In this section we present addition plots that strengthen the results shown in the main paper.

\subsection{Pretrained Baseline}\label{app:pretrained}
Tables \ref{tab:res-overall-en-notune} and \ref{tab:res-overall-langs-notune} show the prediction accuracy of the pretrained models. 

\begin{figure*}[h]
    \centering
    \includegraphics[width=0.75\textwidth]{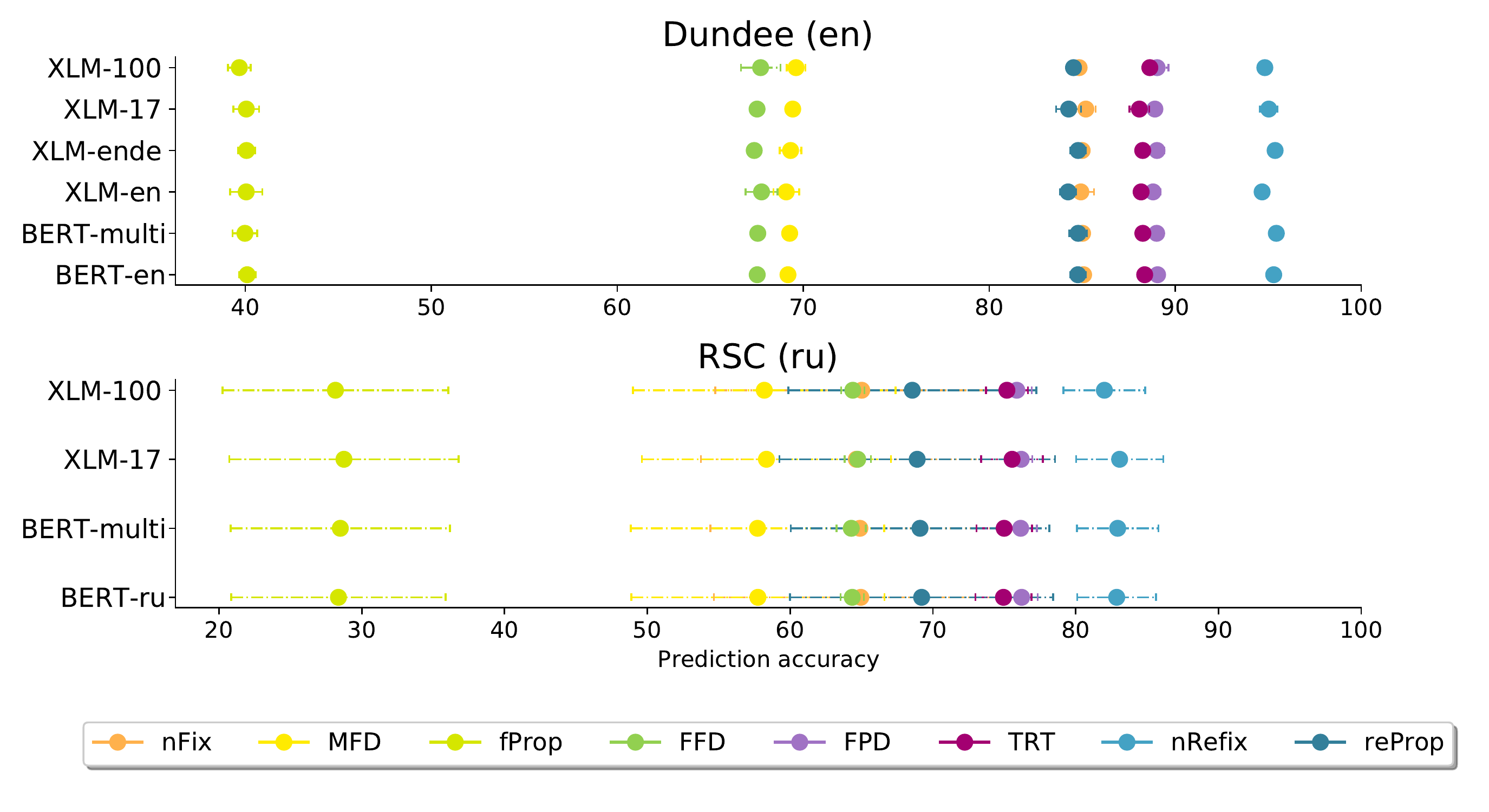}
    \caption{Results of individual gaze features for all pretrained models (without fine-tuning) on the Dundee (en) and RSC (ru) corpora.}
    \label{fig:results-features-notune}
\end{figure*}

\noindent Moreover, Figure \ref{fig:results-features-notune} shows the results of individual gaze features for all pretrained models (without fine-tuning) on the Dundee (en) and RSC (ru) corpora. 

\begin{figure*}[h]
    \centering
    \includegraphics[width=0.65\textwidth]{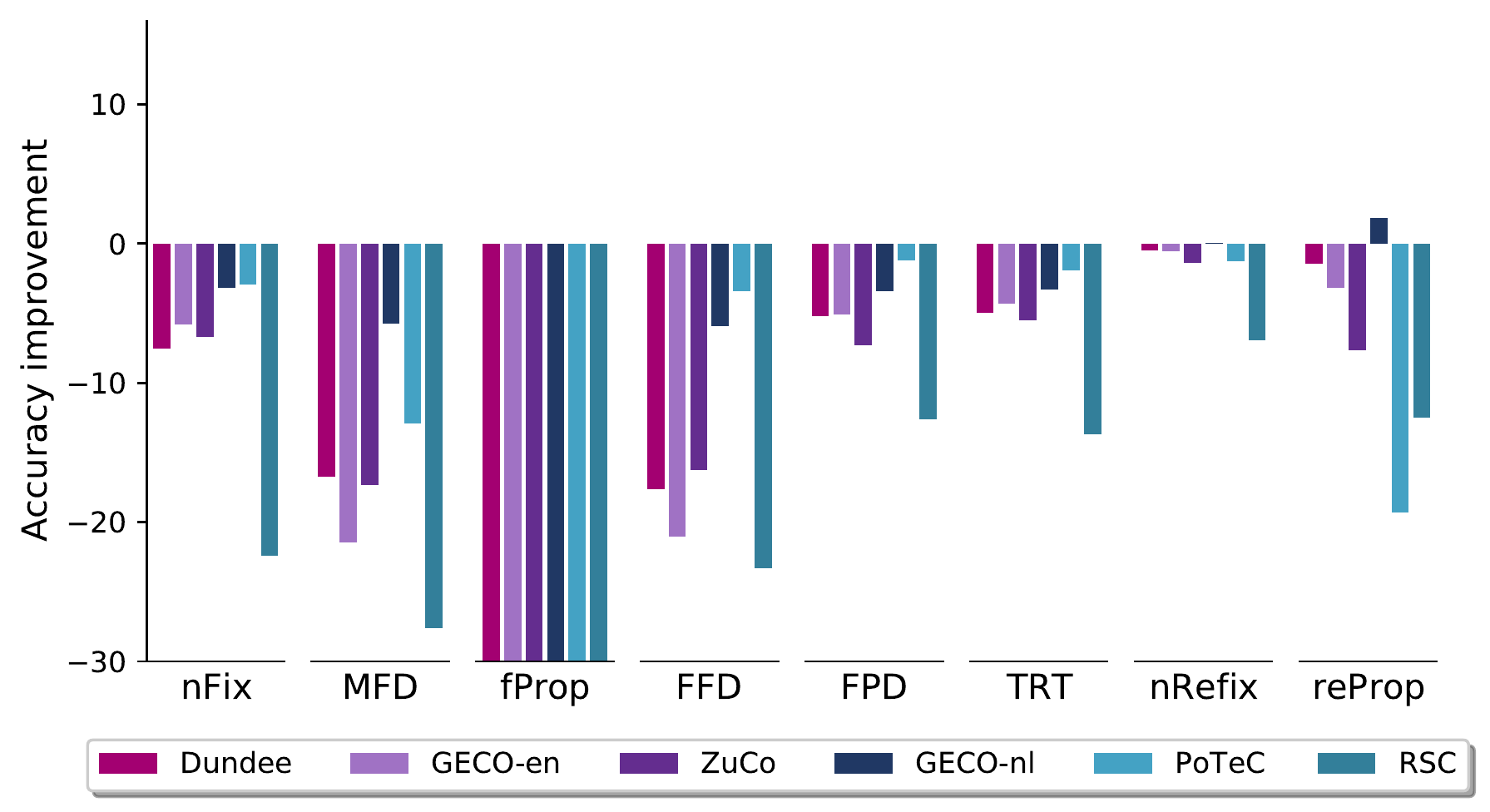}
    \caption{Differences in prediction accuracy for the pretrained \textsc{XLM-100} model predictions (without fine-tuning on eye tracking data) relative to the mean baseline for each eye tracking feature.}
    \label{fig:results-features-langs-selected-notune}
\end{figure*}

\noindent Figure \ref{fig:results-features-langs-selected-notune} presents the differences in prediction accuracy for the pretrained \textsc{xml-100} model predictions relative to the mean baseline for each eye tracking feature. The pretrained models clearly cannot outperform the mean baseline for any language or dataset.

\subsection{Individual Feature Results}\label{app:feat-results}

\begin{figure*}[h]
    \centering
    \includegraphics[width=0.9\textwidth]{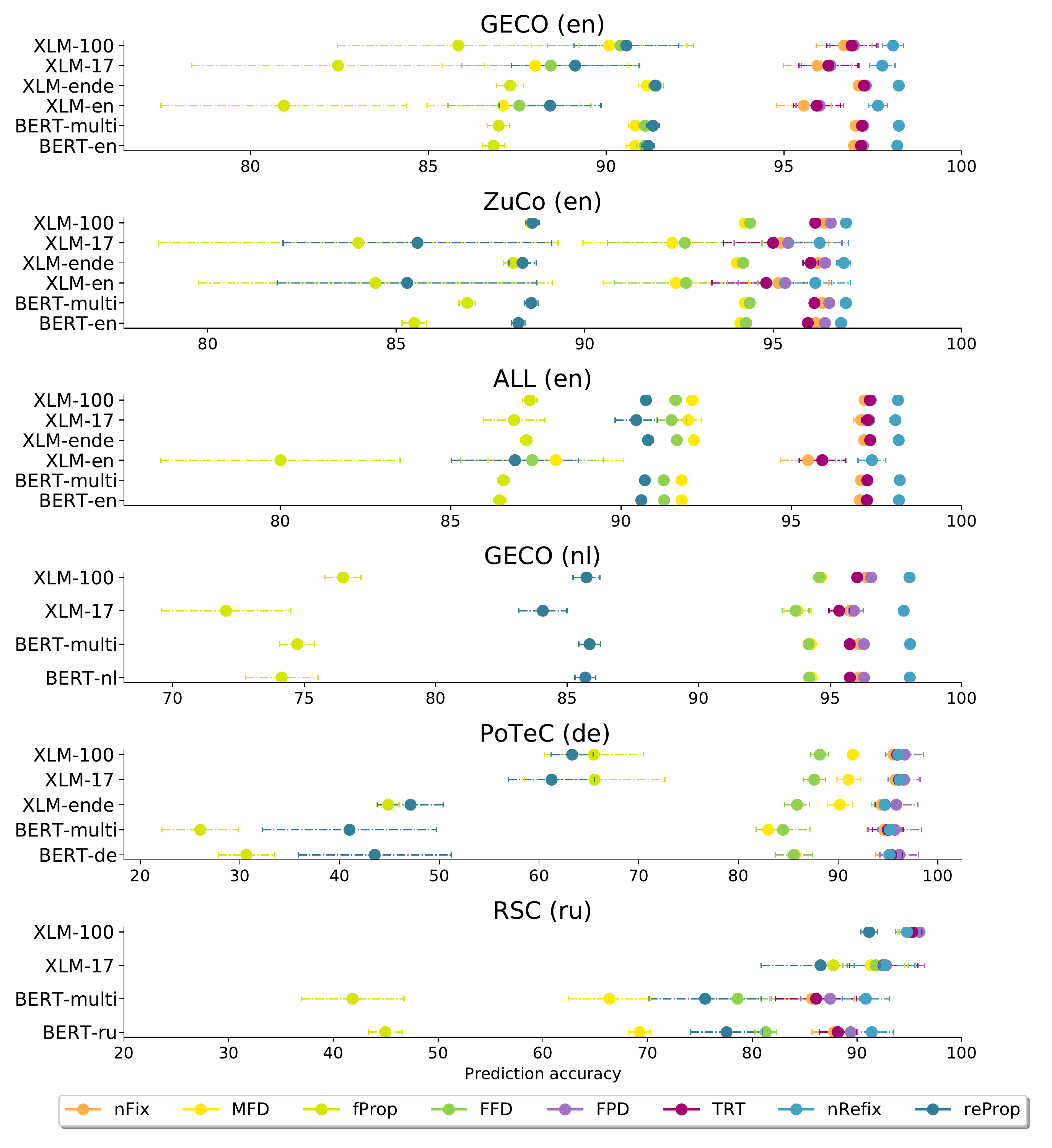}
    \caption{Results of individual eye tracking features for all fine-tuned models on all datasets not presented in the main paper.}
    \label{fig:results-features-rest}
\end{figure*}

Figure \ref{fig:results-features-rest} shows the prediction accuracy of the fine-tuned language models for the individual eye tracking features for all datasets.

\subsection{Word Length Effect}\label{app:word-len}

\begin{figure*}[h]
    \centering
    \includegraphics[width=0.45\textwidth]{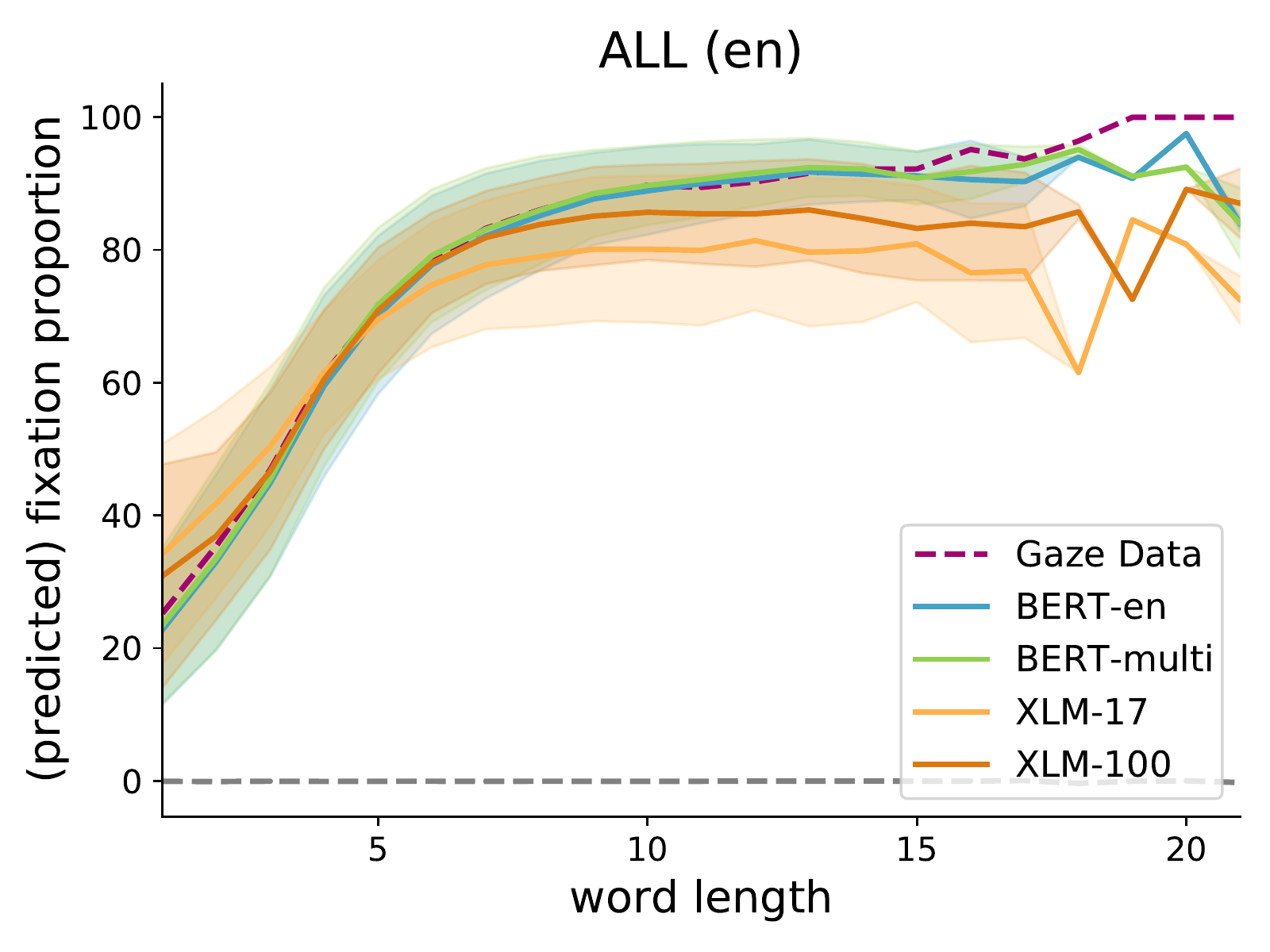}
    \includegraphics[width=0.45\textwidth]{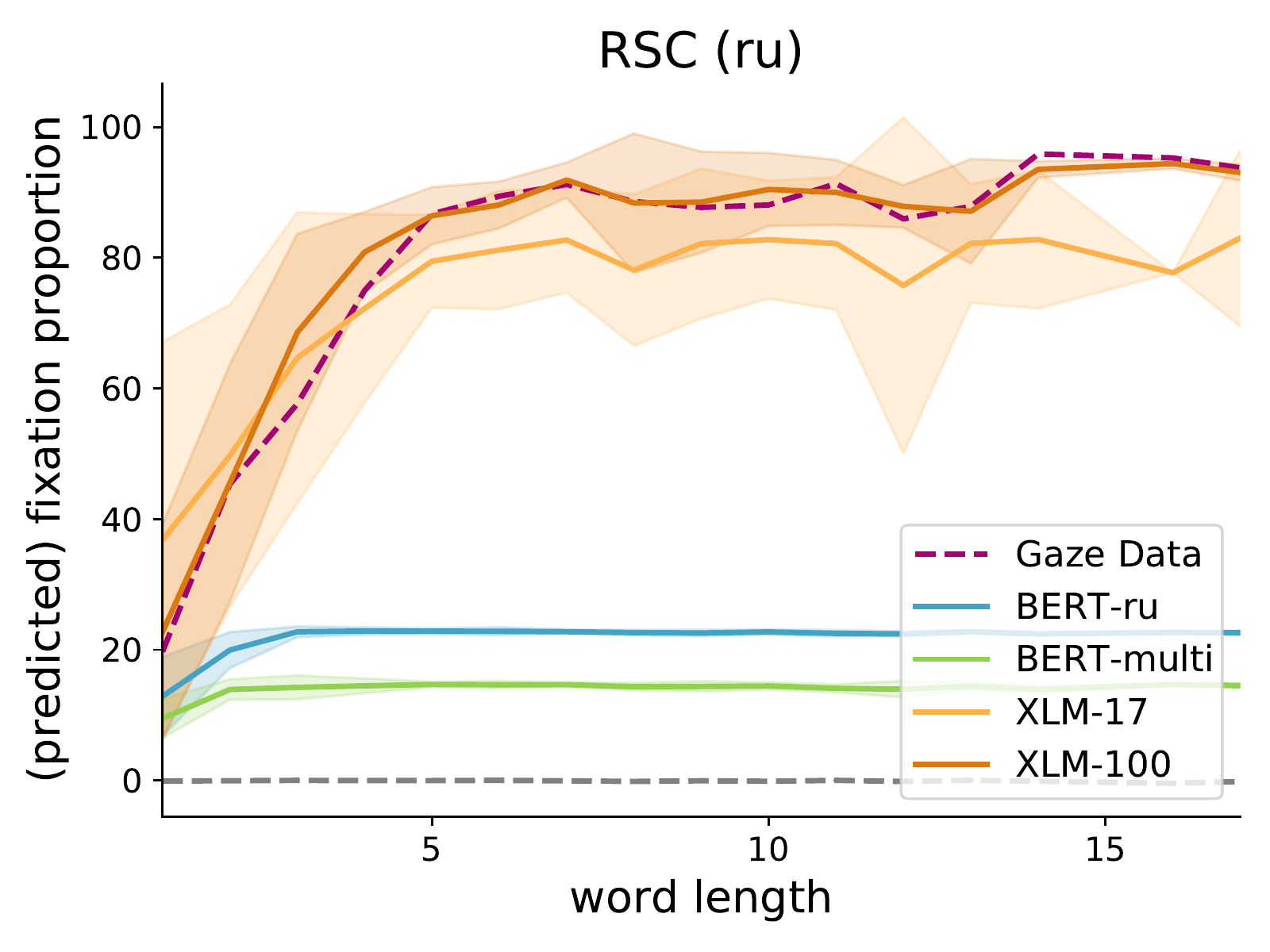}
    \includegraphics[width=0.45\textwidth]{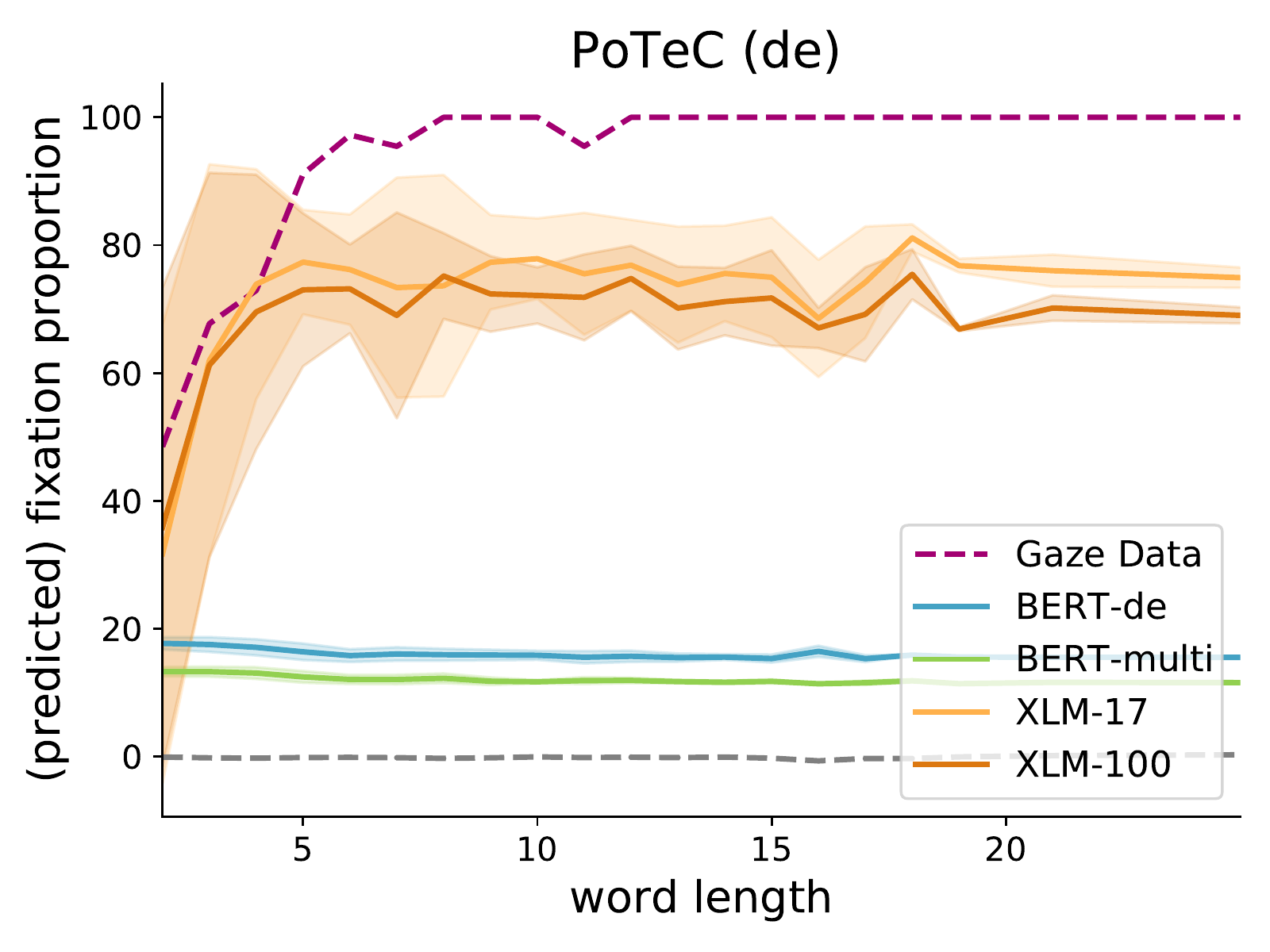}
    \includegraphics[width=0.45\textwidth]{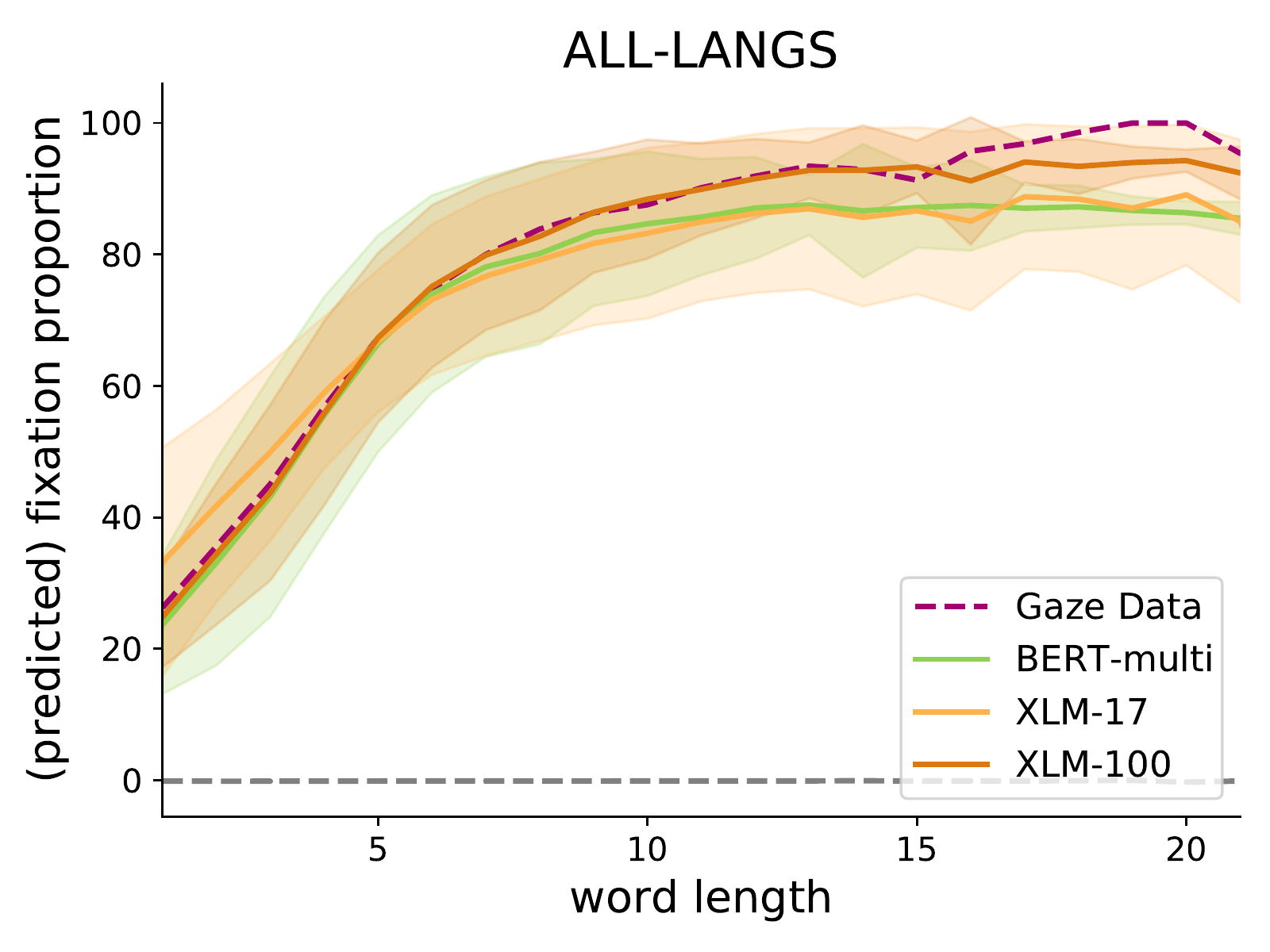}
    \caption{Word length versus predicted fixation probability for Russian, German and English. The gray dashed line is the result of the pretrained \textsc{bert-multi} model without fine-tuning.}
    \label{fig:fix-prob-pred-app}
\end{figure*}

Figure \ref{fig:fix-prob-pred-app} presents the comparison between models predictions and original word length effects for further languages.

\begin{table*}[h]
\centering
\small
\begin{tabular}{lcccccccc}
\toprule
Corpus & \textsc{nFix} & \textsc{MFD} &  \textsc{fProp}   & \textsc{FFD}   & \textsc{FPD} &  \textsc{TRT}  & \textsc{nRefix} & \textsc{reProp}  \\\midrule
Dundee (en) & 0.8 (0.5) & 119.5 (62.1) & 0.6 (0.3) & 120.7 (63.4) & 140.6 (88.5) & 156.1 (105.5) & 0.2 (0.3) & 0.2 (0.2) \\
GECO (en) & 0.8 (0.5) & 128.4 (59.0) & 0.6 (0.2) & 129.3 (60.1) & 143.3 (77.5) & 168.2 (102.4) & 0.2 (0.3) & 0.2 (0.2) \\
ZuCo (en) & 1.1 (0.7) & 78.4 (34.8) & 0.7 (0.3) & 77.3 (34.4) & 92.3 (52.2) & 129.8 (89.7) & 0.4 (0.5) & 0.3 (0.2) \\
GECO (nl) & 0.8 (0.6) & 121.3 (80.1) & 0.6 (0.4) & 121.8 (81.1) & 134.1 (98.0) & 158.1 (131.2) & 0.2 (0.4) & 0.1 (0.2) \\
PoTeC (de) & 2.7 (2.9) & 217.5 (117.3) & 0.8 (0.4) & 167.9 (157.4) & 224.7 (264.2) & 675.6 (727.0) & 1.7 (2.2) & 0.6 (0.5) \\
RSC (ru) & 0.8 (0.4) & 203.4 (115.1) & 0.6 (0.3) & 233.6 (49.5) & 285.1 (101.9) & 314.2 (179.8) & 0.1 (0.1) & 0.1 (0.1) \\
\bottomrule
\end{tabular}
\caption{Mean and standard deviation for all eye tracking features of the corpora used in this work.}
\label{tab:et-features}
\end{table*}

\begin{table*}[h]
\centering
\small
\begin{tabular}{lcccc}
\toprule
\textbf{Model} & \textbf{Dundee} & \textbf{GECO (en)} & \textbf{ZuCo (en)} & \textbf{ALL (en)} \\\midrule
\textsc{bert-en} & 77.42 (0.21) & 77.67 (0.13) & 76.06 (0.38) & 78.69 (0.09) \\
\textsc{bert-multi} & 77.41 (0.21) & 77.68 (0.13) & 76.07 (0.37) & 78.66 (0.07) \\\midrule
\textsc{xlm-en} & 77.21 (0.29) & 77.65 (0.24) & 75.97 (0.60) & 78.47 (0.11) \\
\textsc{xlm-ende} & 77.40 (0.29) & 77.67 (0.10) & 76.10 (0.41) & 78.66 (0.12) \\
\textsc{xlm-17} & 77.31 (0.23) & 77.66 (0.19) & 75.99 (0.39) & 78.39 (0.15) \\
\textsc{xlm-100} & 77.35 (0.29) & 77.63 (0.34) & 75.93 (0.43) & 78.49 (0.11) \\\bottomrule
\end{tabular}
\caption{Prediction accuracy of the pretrained language models aggregated over all eye tracking features for the English corpora, including the concatenated dataset. Standard deviation is reported in parentheses.}
\label{tab:res-overall-en-notune}
\end{table*}

\begin{table*}[h]
\centering
\small
\begin{tabular}{lcccc}
\toprule
\textbf{Model} & \textbf{GECO (nl)} & \textbf{PoTeC (de)} & \textbf{RSC (ru)} & \textbf{ALL-LANGS} \\\midrule
\textsc{bert-nl} & 84.20 (0.10) & - & - & - \\
\textsc{bert-de} & - & 73.55 (3.07) & - & - \\
\textsc{bert-ru} & - & - & 64.83 (2.09) & - \\
\textsc{bert-multi} & 84.28 (0.10) & 73.47 (3.01) & 64.82 (2.11) & 86.22 (0.29) \\\midrule
\textsc{xlm-ende} & - & 73.49 (2.99) & - & - \\
\textsc{xlm-17} & 83.93 (0.16) & 73.17 (2.86) & 65.02 (2.11) & 85.84 (0.27) \\
\textsc{xlm-100} & 83.94 (0.27) & 73.28 (2.91) & 64.67 (2.10) & 85.94 (0.38) \\\bottomrule
\end{tabular}
\caption{Prediction accuracy of the pretrained language models aggregated over all eye tracking features for the Dutch, German and Russian corpora, and for all four languages combined in a single dataset. Standard deviation is reported in parentheses.}
\label{tab:res-overall-langs-notune}
\end{table*}

\end{document}